\documentclass[review]{elsarticle}

\usepackage{lineno}
\usepackage[caption=false,font=normalsize,labelfont=sf,textfont=sf]{subfig}
\usepackage{amsmath}
\usepackage{amssymb}
\usepackage{multicol}
\usepackage{multirow}
\usepackage{enumerate}
\usepackage{color}
\usepackage{graphicx}
\usepackage{threeparttable}
\usepackage{booktabs}
\usepackage{hyperref}
\usepackage[ruled]{algorithm2e}
\SetKwInput{KwNotations}{Notations}

\SetCommentSty{mycommfont}

\newcommand{\maketable}[7]{
\begin{table#4}
    \centering
    \caption{#5}
    \scalebox{#2}{
        \begin{threeparttable}[#3]
            \centering
            {#6}
            {#7}
        \end{threeparttable}
    }
    \label{tbl:#1}
\end{table#4}
}
\modulolinenumbers[5]

\journal{Journal of \LaTeX\ Templates}

\begin{document}

\begin{frontmatter}

    \title{VIBUS: Data-efficient 3D Scene Parsing with \textbf{VI}ewpoint \textbf{B}ottleneck and \textbf{U}ncertainty-\textbf{S}pectrum Modeling}





    \author[thucst,thuair]{Beiwen Tian\corref{intern}}
    \ead{tbw18@mails.tsinghua.edu.cn}

    \author[mcgill,thuair]{Liyi Luo\corref{intern}}
    \ead{liyi.luo@mail.mcgill.ca}

    \author[intel,pku]{Hao Zhao}
    \ead{hao.zhao@intel.com}
    \ead{zhao-hao@pku.edu.cn}

    \author[thuair]{Guyue Zhou\corref{mycorrespondingauthor}}
    \ead{zhouguyue@tsinghua.edu.cn}

    \cortext[intern]{These authors contributed equally to this work.}
    \cortext[mycorrespondingauthor]{Corresponding author}
    \address[thucst]{Department of Computer Science and Technology, Tsinghua University, Beijing, China}
    \address[mcgill]{Bioresource Engineering Department, McGill University, Montreal, Canada}
    \address[intel]{Intel Lab, Beijing, China}
    \address[pku]{Peking University, Beijing, China}
    \address[thuair]{Institute for AI Industry Research, Tsinghua University, Beijing, China}

    \begin{abstract}
        Recently, 3D scenes parsing with deep learning approaches has been a heating topic.
        However, current methods with fully-supervised models require manually annotated point-wise supervision which is extremely user-unfriendly and time-consuming to obtain.
        As such, training 3D scene parsing models with sparse supervision is an intriguing alternative.
        We term this task as data-efficient 3D scene parsing and propose an effective two-stage framework named VIBUS to resolve it by exploiting the enormous unlabeled points.
        In the first stage, we perform self-supervised representation learning on unlabeled points with the proposed Viewpoint Bottleneck loss function.
        The loss function is derived from an information bottleneck objective imposed on scenes under different viewpoints, making the process of representation learning free of degradation and sampling.
        In the second stage, pseudo labels are harvested from the sparse labels based on uncertainty-spectrum modeling.
        By combining data-driven uncertainty measures and 3D mesh spectrum measures (derived from normal directions and geodesic distances), a robust local affinity metric is obtained.
        Finite gamma/beta mixture models are used to decompose category-wise distributions of these measures, leading to automatic selection of thresholds.
        We evaluate VIBUS on the public benchmark ScanNet and achieve state-of-the-art results on both validation set and online test server.
        Ablation studies show that both Viewpoint Bottleneck and uncertainty-spectrum modeling bring significant improvements.
        Codes and models are publicly available at \href{https://github.com/AIR-DISCOVER/VIBUS}{https://github.com/AIR-DISCOVER/VIBUS}.

    \end{abstract}


\end{frontmatter}

\linenumbers

\section{Introduction}

\begin{figure}[t]
    \centerline{\includegraphics[width=0.9\textwidth]{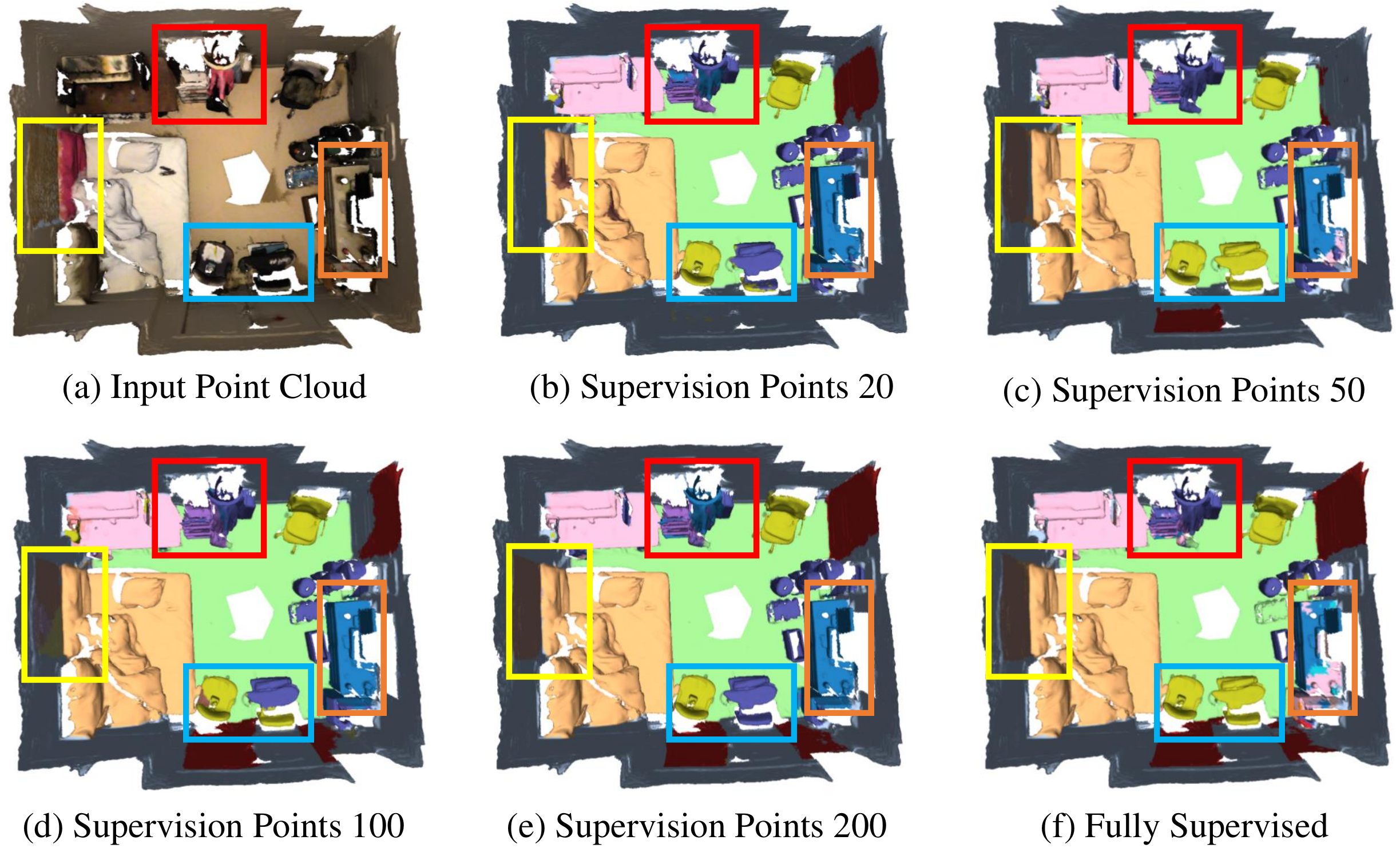}}
    \caption{We address the challenging problem of learning 3D scene parsing models using only semantic annotations on limited points (20 to 200 points per scene).
        (a) shows a point cloud of an indoor scene.
        (b)-(e) demonstrate 3D semantic understanding results using 20 to 200 point annotations per scene.
        (f) shows the result generated by a fully supervised model.
        Except for highlighted regions in boxes, most points are correctly segmented.}
    \label{fig:teaser}
\end{figure}

As a crucial component of 3D computer vision, 3D scene parsing provides the ability of fine-grained understanding for various downstream tasks from robotic manipulation \cite{9561675}, indoor navigation \cite{chaplot2020object} \cite{acharya2019bim} to geographic information system analysis \cite{chen2020method}.
Various deep-learning-based methods, including graph convolutional network \cite{WANG202167}, neural architecture search \cite{LIN2021279} and joint residual-dense optimization \cite{DU202137} have been proposed to address this task in a fully-supervised manner.
However, point-wise annotation of large-scale 3D scenes is very time-consuming \cite{yi2016scalable} \cite{LIN202279}.
Some works propose to address this issue by involving Virtual Reality \cite{zingsheim2021collaborative} \cite{ramirez2020shooting}, Augmented Reality \cite{miksik2015semantic} \cite{shreve2020warhol} or other interactive systems \cite{valentin2015semanticpaint}, in which the procedures for labeling are more user-friendly but still inevitable and time-consuming.
To this end, training 3D scene parsing models with only annotations on sparse points becomes appealing, since annotating all points in a point cloud is not practical in industry.
We call this task data-efficient 3D scene parsing and propose a method named VIBUS to address it.
Fig.~\ref{fig:teaser}-(e) shows the parsing result generated by a model trained using 200 points per scene, which is almost as good as the fully supervised result shown in Fig.~\ref{fig:teaser}-(f).
Since a scene in the ScanNet dataset \cite{dai2017scannet} typically contains more than 50,000 points, a subset of 200 points only amounts to 0.4\% data usage.

The key to data-efficient 3D scene parsing is leveraging numerous unlabeled points (e.g., the other 99.6\% data in the aforementioned case).
There are two natural ideas to achieve this goal: (1) If we can learn meaningful representations from unlabeled points without the usage of semantic annotation, a subsequent data-efficient fine-tuning step is expected to give good results.
This is usually referred to as self-supervised representation learning (SSRL).
(2) Models trained with sparse annotations can generate pseudo labels for unlabeled points in the training set.
If we can harvest correct pseudo labels and fine-tune the model with them, performance is expected to be boosted.
However, applying these two ideas to 3D scenes is challenged by several issues:

\textbf{Challenges on SSRL.}
While exciting progress has been achieved in 2D SSRL \cite{hinton2006reducing}\cite{doersch2015unsupervised}\cite{he2020momentum}, 3D SSRL in point clouds \cite{xie2020pointcontrast}\cite{hou2021exploring} is still an under-explored emerging topic.
Several open problems exist and challenge the effectiveness of existing 3D SSRL methods:
(1) Former 2D SSRL methods treat each image as a sample, whereas 3D SSRL for dense semantic parsing naturally requires us to treat each point as a sample.
This fact leads to an extremely large sample set.
(2) Former contrastive learning methods operate on the sample dimension.
As conceptually shown in the left part of Fig.~\ref{fig:vb_concept}-(b), embeddings for the same sample under different viewpoints are drawn near, and embeddings for different samples are pushed away.
Selecting good sample pairs is an unclear and difficult problem, especially when the 3D sample set is large.
(3) Former contrastive learning methods are troubled with degenerated solutions thus require sophisticated techniques to break the symmetry like weight averaging with momentum \cite{he2020momentum}.
Complex implementation details make them hard to tune.

\textbf{Challenges on pseudo labels.}
The most straightforward way to select pseudo labels is thresholding softmax scores.
However, it is known that softmax scores can be over-confident on false predictions \cite{li2020improving} and vulnerable to adversarial attacks \cite{dong2018boosting}.
Another natural idea is to select pseudo labels that are geometrically close to labeled points, but defining reliable affinity metrics on 3D scene meshes is more difficult than 2D images.
For example, two vertices can be close to each other in the Euclidean space but belong to different categories.
Lastly, no matter which metric is used, tuning thresholds harms the generalization ability of an algorithm.

The proposed method leverages both ideas: an objective called \textbf{VI}ewpoint \textbf{B}ottleneck allows effective 3D SSRL and a module based upon \textbf{U}ncertainty-\textbf{S}pectrum modeling generates reliable pseudo labels.
As such, the method is shortened as VIBUS.
VIBUS addresses aforementioned challenges:
(1) Several recent SSRL methods \cite{zbontar2021barlow}\cite{bardes2021vicreg}\cite{li2021self} switch representation learning from the sample dimension to the feature dimension.
We introduce this idea into 3D SSRL, as conceptually shown in the right part of Fig.~\ref{fig:vb_concept}-(b).
There is no need to select sample pairs or design sophisticated techniques that break symmetry, leading to a simple implementation.
(2) Other than softmax scores, we use Monte Carlo Dropout \cite{gal2016dropout} to generate uncertainty measures that reliably reflect pseudo label quality.
(3) Geodesic distance and normal directions are combined into a unified geometric embedding via spectral analysis, which serves as principled local affinity measures for 3D meshes.
(4) Gamma or beta mixture models are used to decompose uncertainty and spectrum measures so that thresholds can be automatically selected with the mixture distribution.

Our method is empirically successful, achieving state-of-the-art results on the validation set and test server of a widely used benchmark named ScanNet.
Ablations show that both Viewpoint Bottleneck pre-training and uncertainty-spectrum modeling lead to clear performance margins.
Furthermore, we also design an interactive point cloud annotator based on \cite{meshlab} for sparse point cloud annotations. This tool along with our methods have been widely used in an industrial partner for their robot vacuum scene understanding cloud platform.
Our codes and models are publicly available at \href{https://github.com/AIR-DISCOVER/VIBUS}{https://github.com/AIR-DISCOVER/VIBUS}.

\section{Related works}

\begin{figure*}[t]
    \centerline{\includegraphics[width=1\textwidth]{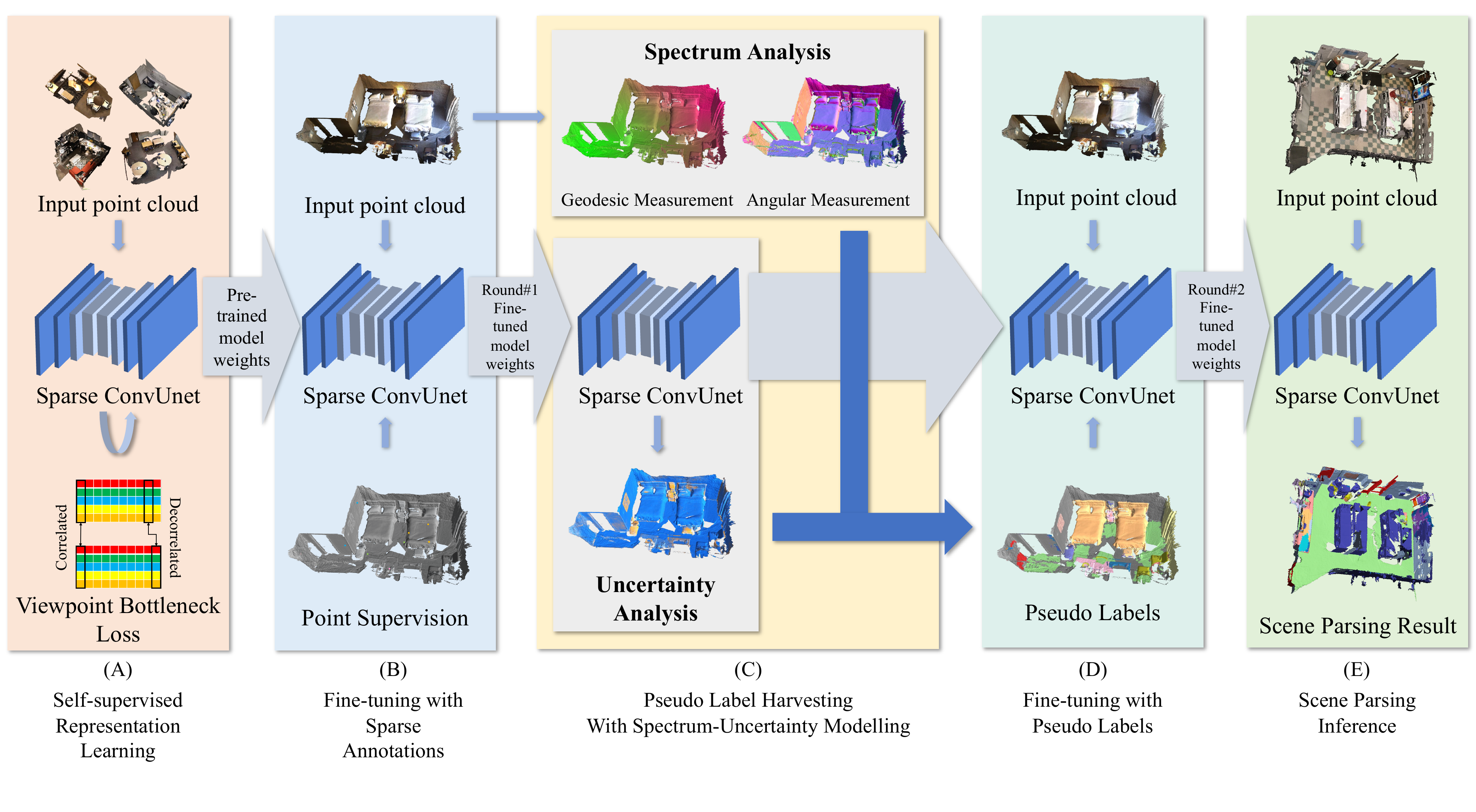}}
    \caption{The overall pipeline of our method.
        In the first stage (A), the backbone network is pre-trained in a self-supervised manner named Viewpoint Bottleneck. With a simple implementation, Viewpoint Bottleneck decorrelates different components of the representations and exploits the intrinsic structures of the unlabeled points.
        In stage (B), the backbone network is fine-tuned with sparse annotations. For each scene, only 20/50/100/200 points have ground-truth labels. With the fine-tuned model, we inference on the training dataset and obtain the predicted label for each point.
        In stage (C), we perform uncertainty analysis and spectrum analysis, and build double mixture models based on the statistics from the analyses. With the mixture models, unreliable labels are filtered out, and reliable labels are selected to form the pseudo label sets.
        In stage (D), The backbone network is fine-tuned once again with the harvested pseudo labels. In stage (E), we inference on the validation set with the fine-tuned model in stage (D).
    }
    \label{fig:main}
\end{figure*}

\subsection{Self-supervised Representation Learning}

SSRL is a special form of unsupervised learning, which aims to learn meaningful representations of the input data without relying on annotations.
The common ground of these methods is to learn representations that are invariant to distortions.
There are several ways to achieve this general principle.
For example, early attempts use different surrogate tasks whose labels can be naturally generated, like pixel value inpainting \cite{pathak2016context}\cite{doersch2015unsupervised}, cross-channel feature regression \cite{zhang2017split}, or rotation prediction \cite{komodakis2018unsupervised}.
Recently, contrastive learning \cite{chopra2005learning}, which was first proposed for metric learning, has drawn significant attention and witnessed great success, such as DenseCL \cite{9578497} and Alonso et al.\cite{9710242}.
However, it is now clear that a trivial contrastive learning formulation suffers from degeneration.
As such, many variants have been proposed to address the issue like weight averaging \cite{he2020momentum}\cite{grill2020bootstrap} or stop gradient \cite{chen2021exploring} among others \cite{van2018representation}\cite{chen2020simple}.

Compared to its 2D counterpart, 3D SSRL is more urgent because of the difficulty of annotating 3D data.
There are already some pioneering works that address the 3D SSRL problem borrowing ideas of contrastive learning, like DepthContrast \cite{9710368}, PointContrast\cite{xie2020pointcontrast} and Contrastive Scene Contexts (CSC)\cite{hou2021exploring}.
DepthContrast works with single-view depth scans that are collected by varied sensors and without any 3D registeration and point correspondences.
PointContrast proposes a PointInfoNCE loss and verifies its effectiveness on a diverse set of scene understanding tasks.
However, it ignores the spatial context around local points, limiting its transferability to complex downstream tasks.
CSC introduces a loss function that contrasts features aggregated in local partitions.
It demonstrates the possibility of using extremely few annotations to achieve high performance on a series of data-efficient settings.
Liu et al. \cite{liu2020p4contrast}\cite{liu2021contrastive} propose to use multi-modal pairs and tuples as elements for RGB-D contrastive learning and scene understanding.
Huang et al. \cite{huang2021spatio} develops a sophisticated spatial-temporal contrastive learning framework to point clouds.

To address the common issues of contrastive learning, a recent SSRL method called Barlow Twins \cite{zbontar2021barlow} proposes an objective function that measures the cross-correlation matrix between the representations of two distorted versions of a sample.
It naturally avoids representation collapse and sample selection since no negative samples are required.
Bardes et al. \cite{bardes2021vicreg} and li et al. \cite{li2021self} enrich this new scheme with a variance term and a principled kernel-based statistical measure.
Inspired by them, we propose the Viewpoint Bottleneck principle, the first non-contrastive 3D SSRL method.

\subsection{Uncertainty analysis in deep learning}
There are many ways to obtain uncertainty measures for deep learning networks.
In our method, we obtain uncertainty measures with multiple inferences with dropout layers \cite{gal2016dropout}.
It officially demonstrates the association between the output variance of multi-layer perceptrons with dropout layers and the predictive distribution.
Zhao et al.\cite{zhao2017pyramid} counts on the dropout layer before the final classifier for fully convolutional models in PSPNet.
In another way, uncertainty measures can be calculated as the root square of the trace of the output covariance matrix.
Jena et al. \cite{jena2019bayesian} exploits the reparameterization trick of adding a Gumbel-distributed random variable onto the activations and dividing them by an annealing parameter.
Lakshminarayanan et al. \cite{lakshminarayanan2017simple} proposed to calculate uncertainty measures by gathering predictions from an ensemble.

\subsection{Spectral clustering}
Spectral clustering is one of the most widely used techniques for data analysis.
Liu et al. \cite{2004Segmentation} firstly applied spectral clustering to 3D mesh segmentation.
However, it is not easy to apply spectral clustering on large-scale point clouds.
The natural way to solve this problem is using sampling methods.
Shinnou et al. \cite{shinnou2008spectral} proposed the Committees-based Spectral Clustering, in which the committees are some points that are near to the cluster centers.
They use these committees to construct the similarity matrix.
Then, the spectral clustering is performed by the reduced similarity matrix.
Chen et al. \cite{chen2011large} developed the Landmark-based Spectral Clustering (LSC), in which they selected $p (\ll n)$ representation data points as the landmarks and represented the original data points as the linear combinations of these landmarks.
Then the spectral embedding of the data can be computed with the landmark-based representation.
In our work, we use the Quadric Error Metrics \cite{garland1997surface} to reduce the size of a large-scale point cloud.
Then, the traditional spectral clustering is applied to the sampling point clouds.

\section{Viewpoint Bottleneck}

\begin{figure}[ht]
    \centerline{\includegraphics[width=0.8\textwidth]{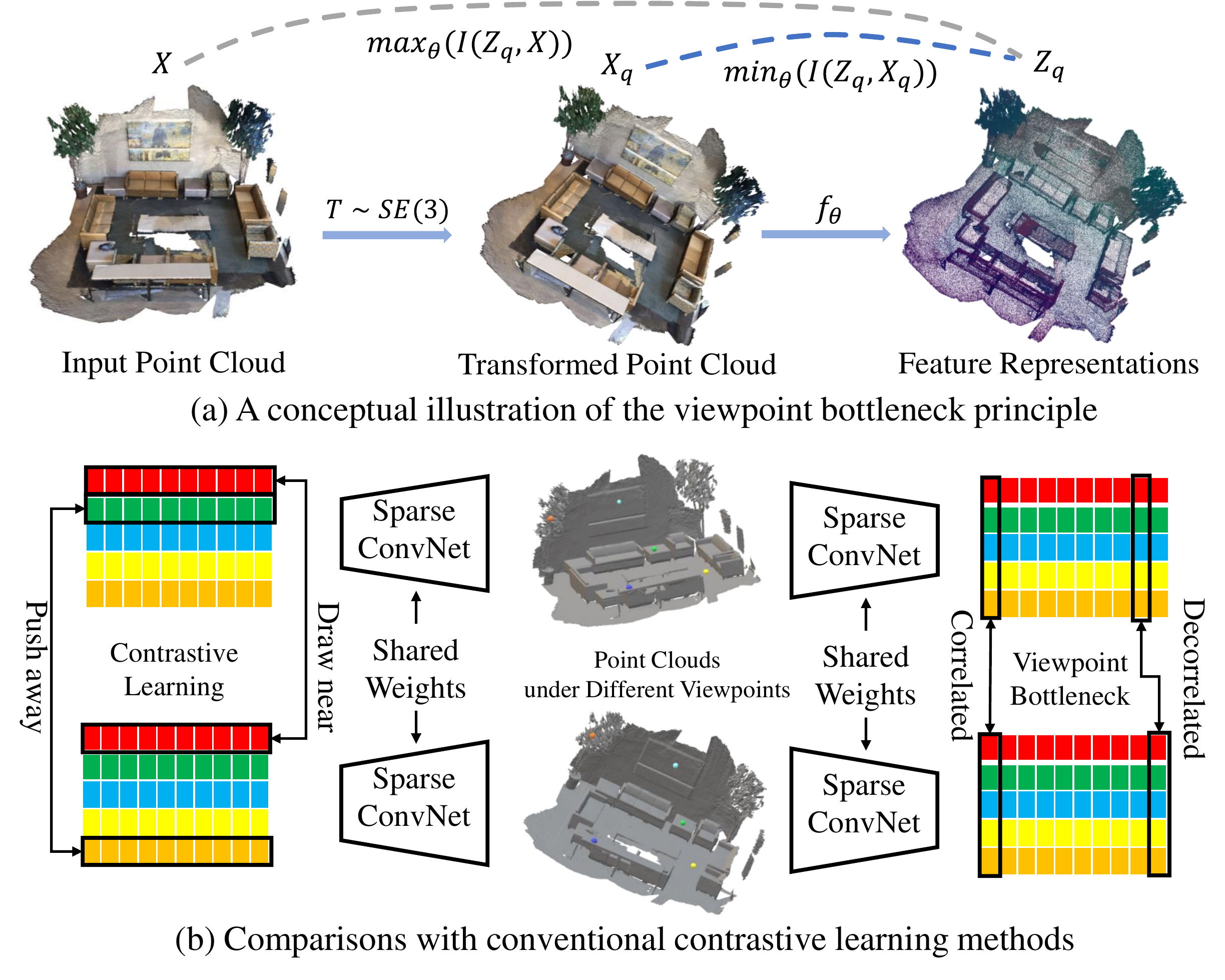}}
    \caption{
        The principles of Viewpoint Bottleneck. In (a), we show that Viewpoint Bottleneck aims to maximize the mutual information between the input point cloud and the feature representation and at the same time minimize the mutual information between the transformed point cloud and the feature representation. In (b), we demonstrate that Viewpoint Bottleneck decorrelates different components of the representations. Instead of operating on the dimension (demonstrated in (b)-left) like former contrastive learning methods, Viewpoint Bottleneck operates on the feature dimension (demonstrated in (b)-right), leading to a simple pipeline free of degradation and sampling.
    }
    \label{fig:vb_concept}
\end{figure}

In this section, we describe the details of the proposed self-supervised representation learning framework named Viewpoint Bottleneck.
By introducing the novel Viewpoint Bottleneck loss, we aim to exploit unlabeled points and learn meaningful representations.

\subsection{Self-supervised Representation Learning Pipeline }
In this section, we describe the general pipeline of Viewpoint Bottleneck.

Firstly, a point cloud $X$ is transformed into two viewpoints by random geometric transformations.
Transformed 3D point clouds are denoted by $X_p$ and $X_q$, respectively.
Then, the two randomly transformed point clouds are fed into the same sparse convolution network $f_\theta$ to obtain two high-dimensional representations $Z_p = f_\theta(X_p)$ and $Z_q = f_\theta(X_q)$.

Next, we sample the representations by Farthest Point Sampling (FPS) from $M_i\times D$ to $H_i\times D$, where $D$ is the dimension of the representations and $M_i, H_i$ are the numbers of points before and after sampling.
FPS starts with a random point as the first proposal candidate and iteratively selects the farthest point from the already selected points until $H_i$ candidates are selected.
We set $H_i\ll M_i$ to keep computation tractable.
In the meantime, FPS guarantees that the sampled subset is a reasonable abstraction of the original point cloud.

After FPS, the cross-correlation matrix $\mathcal{Z} (D\times D)$ is computed on the batch dimension, from the two sampled representations $Z_{p}'$ and $Z_{q}'$.
Finally, the network is trained with the proposed Viewpoint Bottleneck loss, as detailed later. 

\subsection{Viewpoint Bottleneck Loss}

Inspired by Barlow Twins \cite{zbontar2021barlow}, we derive our Viewpoint Bottleneck objective from an information theory perspective \cite{2015Deep}\cite{2000The}.
As shown in Fig. \ref{fig:vb_concept}, the information bottleneck principle assumes that meaningful representations should reserve as much information as possible about the input while affected by random viewpoint transformations as little as possible.
Formally the objective is stated as:
\begin{equation}
    {\rm VB}_\theta\triangleq{I(C_\theta,B)-\beta I(C_\theta,A)}\label{eq:IB}
\end{equation}
where $I(\cdot,\cdot)$ is mutual information between two random vectors.
Here $A$, $B$, and $C_\theta$ correspond to input point clouds, randomly transformed point clouds, and learned representations, respectively.
$\beta$ is a balancing weight.
By definition, the mutual information can be calculated as $I\left(C_\theta,B\right)=H\left(C_\theta\right)-H\left(C_\theta\middle|B\right)$ (similarly for $I\left(C_\theta,A\right)$).
$H\left(C_\theta\right)$ is the entropy of $C_\theta$ while $H\left(C_\theta\middle|B\right)$ is the conditional entropy of $C_\theta$ given $B$.
As the function $f_\theta$ is deterministic, $H(C_\theta|B)$ cancels to 0.
We rewrite the Eq.\ref{eq:IB} as:
\begin{equation}
    {\rm VB}_\theta=H\left(C_\theta\right)-\beta\left[H\left(C_\theta\right)-H\left(C_\theta\middle|A\right)\right]\label{eq:entropy}
\end{equation}

By dividing Eq.\ref{eq:entropy} with $\beta$, we have:
\begin{equation}
    {\rm VB}_\theta=H\left(C_\theta\middle|A\right)+\frac{1-\beta}{\beta}H\left(C_\theta\right)\label{eq:ibp}
\end{equation}

However, the optimization of ${\rm VB}_\theta$ becomes a challenge as evaluating the entropy of a generic random vector is intractable. We hereby derive a surrogate loss. We assume the representation $\vec{C_\theta}$ is distributed as a Gaussian:
$\vec{C_\theta} = \left(c_{\rm \theta 1},c_{\rm \theta 2},\cdots,c_{\rm \theta D}\right) \sim\mathcal{N}_D(\mu,\Sigma_{C_\theta})$, where $D$ is the dimension of the representation.

$\Sigma_{C_\theta}$ can be decomposed into $\Sigma_{C_\theta}\ =\ E^TE$, and $Y=E^{-1}({\vec{C_\theta}-\vec{\mu}})\sim\mathcal{N}\left(\vec{0},I\right)$. In the following derivation, $|\cdot|$ means taking determinant. The probability density function of $\vec{C_\theta}$ is:
\begin{equation}
    \begin{split}
        &p\left(c_{\rm \theta 1},c_{\rm \theta 2},\cdots,c_{\rm \theta D}\right)\\
        &=\frac{1}{\left(2\pi\right)^\frac{D}{2}\left|\Sigma_{C_\theta}\right|^\frac{1}{2}}e^{-\frac{1}{2}\left[\left({\vec{C_\theta}-\vec{\mu}}\right)^T\Sigma_{C_\theta}^{-1}\left(\vec{C_\theta}-\vec{\mu}\right)\right]}
    \end{split}
\end{equation}
where $D$ is the dimension of $\vec{C_\theta}$.

Then according to \cite{cai2015law}:
\begin{equation}
    \begin{split}
        H\left(C_\theta\right)&=-\int{p\left(c_\theta\right)\log{p\left(c_\theta\right)dc_\theta}}\\
        &=-\int{p\left(c_\theta\right)\log{\frac{e^{-\frac{1}{2}\left({\vec{C_\theta}-\vec{\mu}}\right)^T\Sigma_{C_\theta}^{-1}\left(\vec{C_\theta}-\vec{\mu}\right)}}{\left(2\pi\right)^\frac{D}{2}\left|\Sigma_{C_\theta}\right|^\frac{1}{2}}}dc_\theta}\\
        &=-\int{p\left(c_\theta\right)\log{\frac{1}{\left(2\pi\right)^\frac{D}{2}\left|\Sigma_{C_\theta}\right|^\frac{1}{2}}}}dc_\theta\\
        &\quad-{\int{p\left(c_\theta\right)\log{e^{-\frac{1}{2}\left({\vec{C_\theta}-\vec{\mu}}\right)^T\left(E^{-1}\right)^{T}E^{-1}\left(\vec{C_\theta}-\vec{\mu}\right)}}dc_\theta}}\\
        &=\log{\left(2\pi\right)^\frac{D}{2}\left|\Sigma_{C_\theta}\right|^\frac{1}{2}}-\int{p\left(Y\right)\log{e^{-\frac{1}{2}Y^TY}}dY}\\
        &=\log{\left(2\pi\right)^\frac{D}{2}\left|\Sigma_{C_\theta}\right|^\frac{1}{2}}-\sum_{i=1}^{D}\log{e^{-\frac{1}{2}\left|y_i^2\right|}}\\
        &=\log{\left(2\pi\right)^\frac{D}{2}\left|\Sigma_{C_\theta}\right|^\frac{1}{2}}+\log{e^\frac{D}{2}}\\
        &=\frac{1}{2}\log{{(2 \pi e)}^D}\left|\Sigma_{C_\theta}\right|
    \end{split}
\end{equation}

\begin{equation}
    \begin{split}
        H\left(C_\theta\middle|A\right) &=\sum_{a} p\left(a\right)H\left(C_\theta\middle|A=a\right)\\
        &=\frac{1}{2}\mathbb{E}_A\log(2 \pi e)^{n}|\Sigma_{C_{\theta}|A}|
    \end{split}
\end{equation}

As such the Viewpoint Bottleneck objective Eq.\ref{eq:ibp} is turned into a practical surrogate loss function as:
\begin{equation}
    {\rm VB}_\theta=\mathbb{E}_A\log{\left|\Sigma_{C_\theta|A}\right|}+\frac{1-\beta}{\beta}\log{\left|\Sigma_{C_\theta}\right|}\label{eq:ibf}
\end{equation}

Finally, we make several more simplifications:
1) Since Eq.\ref{eq:ibf} is only meaningful when $\beta > 1$, we use a positive constant $\lambda$ to replace $\frac{\beta-1}{\beta}$.
2) In the second term of the Eq.\ref{eq:ibf} we use the Frobenius norm of $\Sigma_{C_\theta}$ as the metric.
3) Recall that the first term of Eq.\ref{eq:ibf} is derived from the conditonal entropy of the representation $C_\theta$ given $A$. Minimizing it is equivalent to maximizing the diagonal elements of $\mathcal{Z}$ (the aforementioned $D \times D$ cross-correlation matrix).

\subsection{Viewpoint Bottleneck: Implementation}
After making aforementioned three simplifications to Eq.\ref{eq:ibf}, we obtain the final Viewpoint Bottleneck loss function as:
\begin{equation}
    \mathcal{L}_{\rm VB}\triangleq \|\Gamma_{\lambda}(\mathcal{Z})-\mathcal{I}\|_F \label{eq:vb}
\end{equation}
where $\mathcal{I}$ denotes the identity matrix, $\Gamma_{\lambda}(\cdot)$ denotes scaling off-diagonal elements by positive constant $\lambda$, and $\mathcal{Z}$ is the cross-correlation matrix computed between the two representations of different viewpoints along the feature dimension:

\begin{equation}
    \mathcal{Z} \triangleq \left(\widetilde{Z_p'}\right)^T\widetilde{Z_q'}\label{eq_cc}\\
\end{equation}

$\widetilde{Z'}$($H\times D$) is the representation obtained by normalizing $Z'$ along the sample dimension.  

$\mathcal{L}_{\rm VB}$ aims to push the diagonal elements of $\mathcal{Z}$ towards 1, so that the representations are invariant to random geometric transformations. Meanwhile, by forcing the off-diagonal elements towards 0, we decorrelate different vector components of the representations that accordingly contain non-redundant information about the original point cloud.

Without complex network structures (e.g., asymmetric network, weight averaging), the intrinsic structures between enormous unlabeled points are fully leveraged by the training scheme of Viewpoint Bottleneck, making it an efficient way to learn effective representations from the unlabeled points.

The pipeline of Viewpoint Bottleneck is also demonstrated in Algorithm. \ref{algorithm:1}.
We then fine-tune the network using labels and pseudo labels harvested by the methods that we introduce in Section \ref{sec: pseudo label harvesting}.


    {\centering
        \begin{minipage}{.9\linewidth}
            \begin{algorithm}[H]
                \caption{3D SSRL with Viewpoint Bottleneck}\label{algorithm:1}
                \KwNotations{
                    \begin{itemize}
                        \setlength\itemsep{0em}
                        \item $\mathbb{X}$: a set of 3D point clouds;
                        \item $\mathcal{T}$: the distribution of geometric transformations;
                        \item $f_\theta$: sparse ConvUNet parameterized by $\theta$;
                        \item $K$: the number of optimization steps;
                        \item $\lambda$: the control parameter in VB loss
                        \item ${\rm FPS}$: Farthest Point Sampling, takes a poing cloud as input \\ and outputs the indices of sampled points
                    \end{itemize}}
                \For{$k=1$ \KwTo $K$}{

                \tcc{sample transformations}
                $T_{p},T_{q} \sim \mathcal{T}$

                \tcc{take an input point cloud}
                $X \in \mathbb{X}$

                \tcc{apply transformations}
                $X_{p}=T_{p}(X), \  X_{q}=T_{q}(X)$

                \tcc{extract features}
                $Z_{p}={f}_{\theta_k}(X_{p}), \  Z_{q}={f}_{\theta_k}(X_{q})$

                \tcc{perform farthest point sampling }
                \tcc{on the original point clouds}
                \tcc{use the sampled indices to select features}
                $Z_{p}' = Z_{p}[{\rm FPS(X_{p})}], \ Z_{q}' = Z_{q}[{\rm FPS(X_{q})}]$

                \tcc{calculate cross-correlation}
                $\mathcal{Z} = (Z_{p}')^T Z_{q}'$

                \tcc{calculate VB loss}
                $\mathcal{L}_{\rm VB}= \|\Gamma_{\lambda}(\mathcal{Z})-\mathcal{I}\|_F$

                \tcc{optimize network parameters}

                $\theta_{k+1}={\rm optim}(\theta_{k},\mathcal{L}_{\rm VB})$

                }
            \end{algorithm}
        \end{minipage}
        \par
    }




\section{Uncertainty-Spectrum Analysis and Pseudo Label Harvesting}
\label{sec: pseudo label harvesting}

\begin{figure*}[t]
    \centerline{\includegraphics[width=1\textwidth]{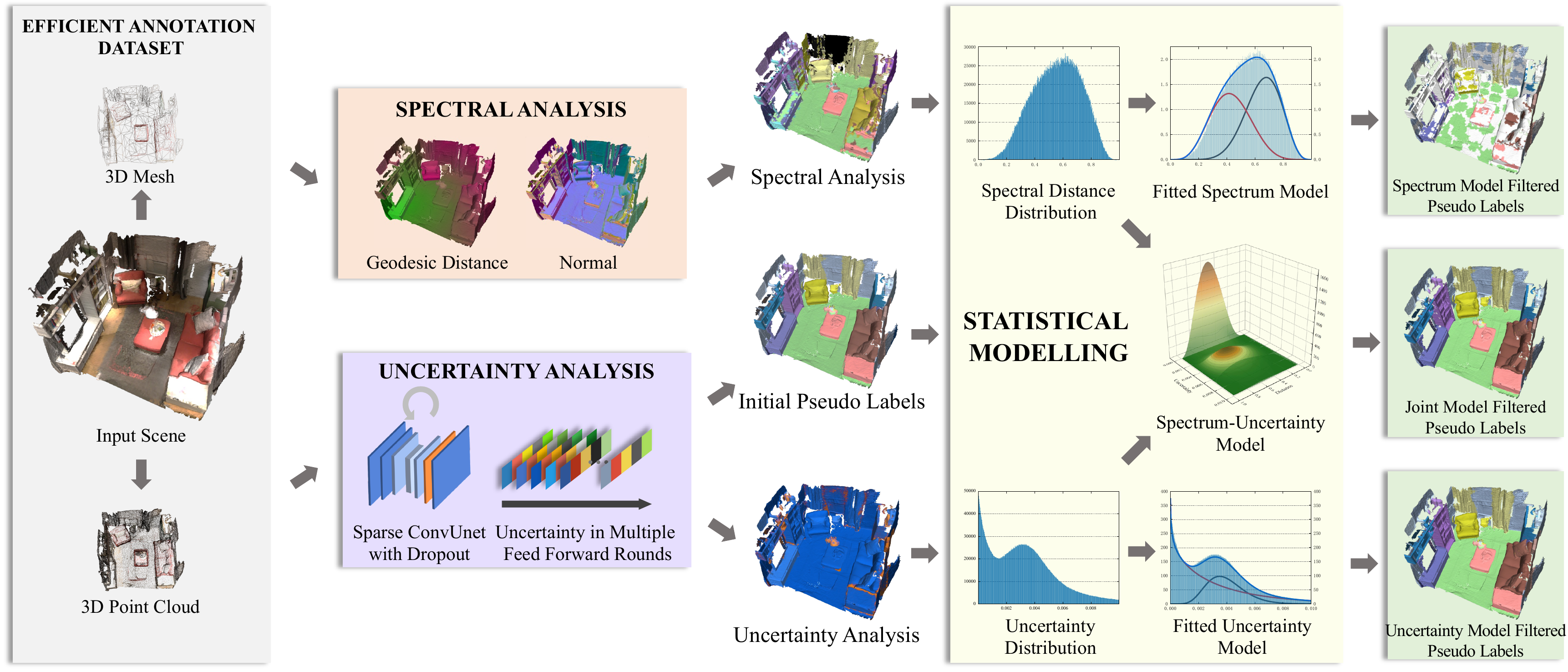}}
    \caption{
        Illustration of the harvesting process of pseudo labels.
        To perform uncertainty analysis, we repeatedly inference with the pre-trained model with effective dropout layers and calculate the variances of the digits as the uncertainty of the predicted labels.
        To perform spectral analysis, we perform k-means clustering with the combination of geodesic and angular distance as the distance measure. The spectrum distance of a point to its cluster center is regarded as the spectrum distance of the associating predicted label.
        With statistics obtained from the analyses, we fit double mixture models for each category. Each double mixture model consists of a reliable and an unreliable component. The pseudo labels set are then generated by keeping the predicted labels that have a larger probability to be drawn from the reliable part.
    }
    \label{fig:specunc}
\end{figure*}

To address the issue of pseudo label harvesting, we seek to fully exploit the labeled points to generate new labels for the unlabeled points.
To achieve this goal, we first fine-tune the pre-trained model with the sparse annotations and obtain the predictions of categories for each point in the scene.
Then the predictions are then filtered so that predictions with high confidence are collected as pseudo labels, which we regard as the supervision for the second fine-tuning process.

In the following sections, we propose three strategies for the aforementioned filtering process: uncertainty modeling, spectrum modeling, and uncertainty-spectrum joint modeling.

\subsection{Uncertainty Modeling}
\label{sec: unc analysis}

When a neural network generates deterministic predictions, it can be problematic that the model is overconfident with wrong predictions, since deterministic results cannot show the model confidence.
To alleviate this situation, a better practice is to generate the predictions with probability distributions.

One possible approach is to treat the network parameters as random variables, but iterating all possible parameters of a neural network is unrealistic.
Gal et al. \cite{gal2016dropout} put forward a workaround to assess the uncertainty of deep neural networks by running the model with functional dropout layers for multiple rounds and calculating the variances of the logits.
Recent work by Zhao et al. \cite{zhao2020pointly} argues that dropout-based uncertainty is a valid criterion for label quality and achieves state-of-the-art performances of semantic segmentation on real-world 2D image datasets by applying the criterion to select reliable pseudo labels.

Following this insight, we aim to leverage the class-wise dropout-based uncertainty of pseudo labels in 3D scenes, which was not well studied before this work.
With dropout layers functioning in the pre-trained network, we run inferences on the training set multiple times and calculate the variance of digits in multiple runs as the uncertainty of the predictions.

\begin{figure}[ht]
    \centering
    \includegraphics[width=0.95\textwidth]{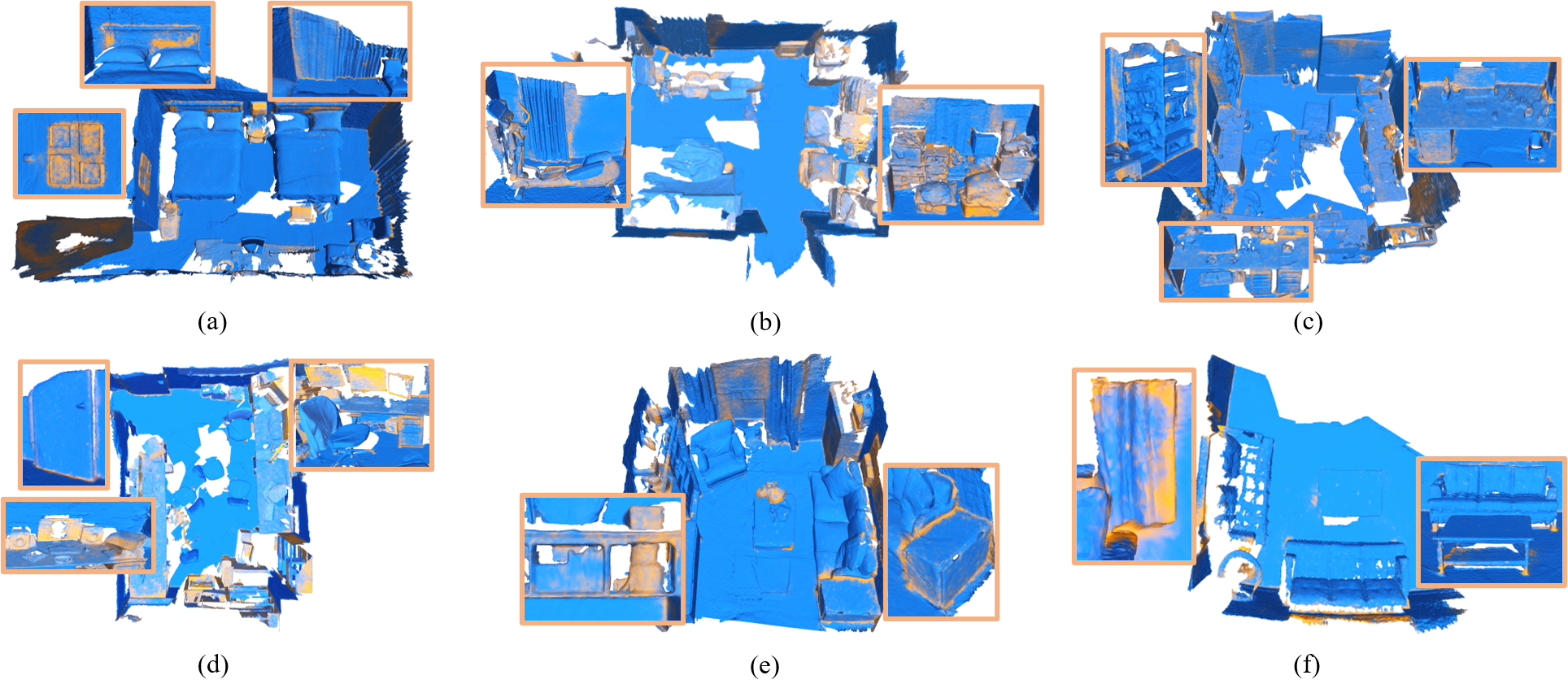}
    \caption{The visualization of uncertainty. Yellow parts have relatively higher uncertainty than blue parts.}
    \label{fig:unc_analysis}
\end{figure}

\begin{figure}[ht]
    \centerline{\includegraphics[width=0.95\textwidth]{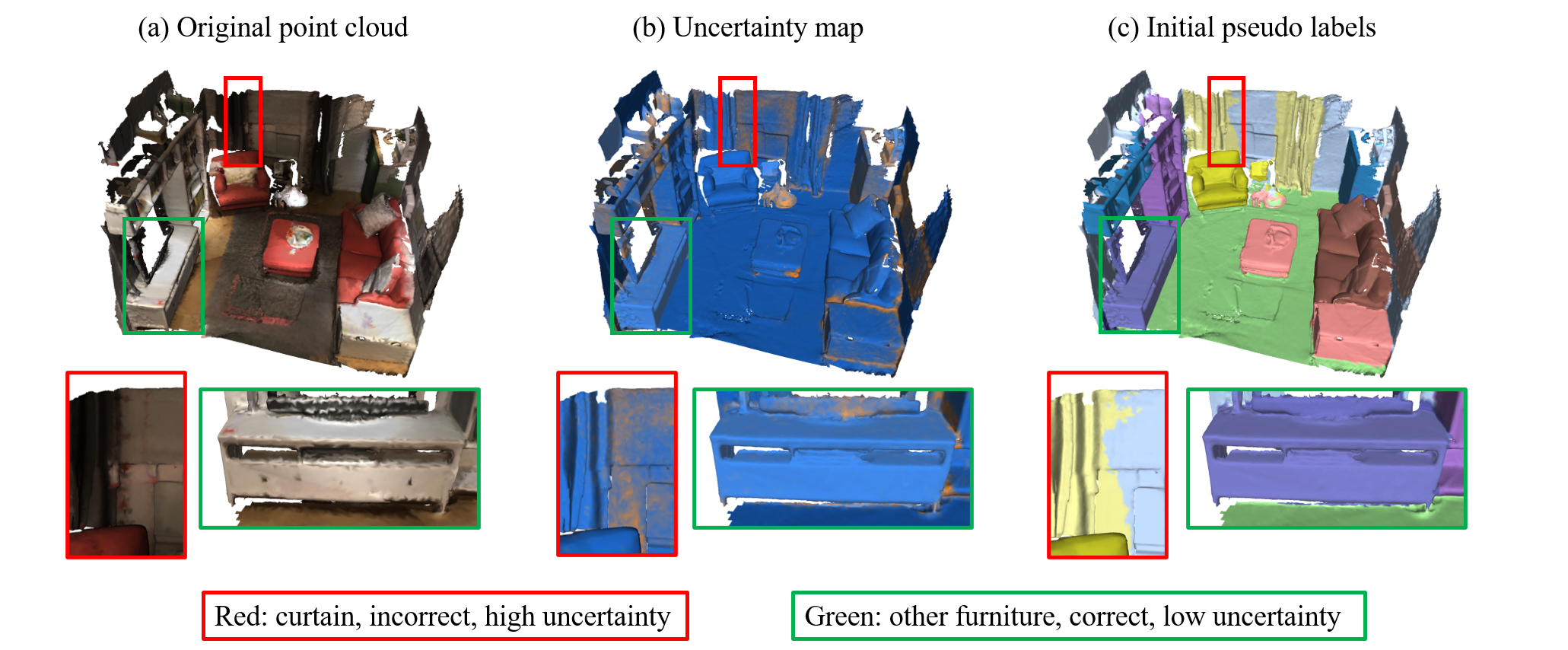}}
    \caption{Red boxes: wrong pseudo labels with high uncertainty.
        Green boxes: right pseudo labels with low uncertainty.}
    \label{fig:uncertainty}
\end{figure}

We show visualized uncertainty in Fig. \ref{fig:unc_analysis}.
We observe that points with incorrect predictions tend to have higher uncertainty and vice versa, as shown in Fig. \ref{fig:uncertainty}.
Furthermore, the frequency histograms show non-trivial patterns that are similar with \cite{zhao2020pointly}.

Following this keen observation and also inspired by \cite{zhao2020pointly}, we propose a double mixture model to fit the distribution of uncertainty in a class-wise manner.
We assume that the uncertainty distribution consists of a certain and an uncertain part and use the two components of the mixture model to fit these two parts, respectively.
We adopt gamma distribution as the probability density function, since the uncertainty is intrinsically variance thus always positive and the distribution is long-tailed in our observations.
Therefore, the mathematical formulation of the probability density of the uncertainty being $x$ is given by
\begin{equation}
    \begin{aligned}
        P(x|\theta_1, \theta_2) & = \omega_1\cdot P(x|\theta_1) + \omega_2\cdot P(x|\theta_2)       \\
        P(x|\theta)             & = \frac{b^{a}x^{a-1}{\rm e}^{-bx}}{\Gamma(a)},\ \theta = \{a, b\} \\
    \end{aligned}
\end{equation}
where $\omega_{1}$ is the probability that a sample is drawn from the certain part and $\omega_{2}$ is the probability that a sample is drawn from the uncertain part.
$\theta_1$ and $\theta_2$ are the parameter sets of the two components.
Each parameter set $\theta$ consists of two independent parameters $a$ and $b$ while $a$ is the shape parameter that controls the general shape of the distribution and $b$ is the scale parameter demonstrating the degree of concentration.

The detailed math derivation of fitting double mixture models is described in Sec. \ref{sec: double model fitting}.
With the double mixture model, we select from the generated labels and harvest reliable pseudo labels that are more likely to belong to the certain part than the uncertain part.

\subsection{Spectrum Modeling}
\label{sec: spec clustering}

Intuitively, local areas around points with ground-truth labels tend to share the same semantic labels.
However, in uncertainty modeling, this empirical fact is overlooked.
In this section, we seek to apply clustering methods on the dataset to further utilize the locality of semantic labels.
To fully exploit the mesh connectivity information in the dataset, we choose to perform clustering based on spectrum distances rather than Euclidean distances, since spectral analysis is able to entangle different measures into a uniform feature space.

\subsubsection{Distance Metric}

To perform spectral clustering, the metric to evaluate the locality of points is crucial, and spatial distance between points is a crucial measurement of the locality.

However, Euclidean distance fails to play this role as adjacent points in Euclidean space may differ dramatically in semantics due to the absence of mesh edge connection.
To fully exploit the relations between points from mesh edges, we adopt geodesic distance to measure the spatial distance between points.

\begin{figure}[ht]
    \centerline{\includegraphics[width=0.95\textwidth]{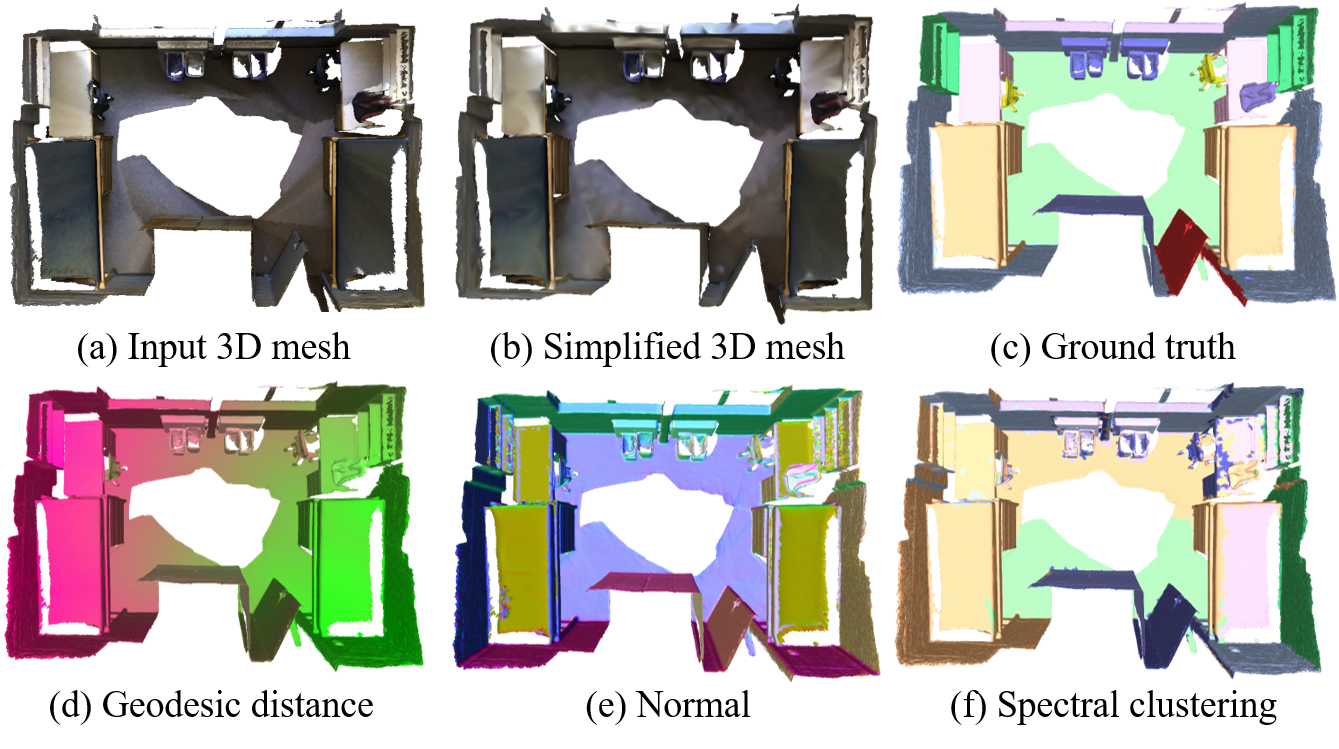}}
    \caption{The visualization of spectral analysis. (a) An input 3D mesh (b) The 3D mesh simplified with $Quadric\ Edge\ Collapse\ Decimation$ \cite{garland1997surface} \cite{muntoni2021pymeshlab}. (c) The ground truth semantic visualization of all vertices. (d) Visualization of geodesic distances from a vertex to the other vertices in the scene. (e) Visualization of normal vectors of the vertices. (f) Visualization of the clustering results.}
    \label{fig: spectral}
\end{figure}

Instead of calculating in a brute-force manner, we estimate geodesic distances along mesh edges by heat diffusion method \cite{crane2013geodesics} which is much faster and reasonably accurate.
In this method, the shortest distances along meshes between one point $v$ and all other points are estimated by simulation of heat diffusion along mesh edges starting at $v$.
Specifically, the geodesic distance between point $x$ and $y$ is derived as:
\begin{equation}
    \label{eq:geo}
    \begin{aligned}
        d_g(x, y)=\lim _{t \rightarrow 0} \sqrt{-4 t \log h_{t, x}(y)}
    \end{aligned}
\end{equation}
where $h_{t, x}(y)$ is the heat kernel measuring the number of particles transferred from a source point $x$ to a target point $y$ after a short period $t$.
Fig. \ref{fig: spectral}(d) illustrates the geodesic distances between one point and all other points in a 3D scene.

On the other hand, points that belong to orthogonal planes tend to have inconsistent semantic labels, thus low locality.
Points near plane intersections may have different semantic labels. However, with the geodesic metric, they are likely to be grouped into the same cluster due to the small geodesic distances.
To address this issue, we use the angular distance $d_a$ as an additional metric to measure the extent to which the spatial neighborhoods of points are aligned.

To calculate the angular distances between two points, we first fit two closest planes of the two points and their 10 nearest neighbors.
The normal vectors of the two planes are considered as the normal vectors of the two points. We then calculate the angular distance between the two points as:
\begin{equation}
    d_a(x, y)= 1 - |\vec{n_{x}} \cdot \vec{n_{y}}|
\end{equation}
where $\vec{n_x}$ and $\vec{n_y}$ denote the normal vector of two points.
Fig.\ref{fig: spectral})(e) illustrates the normal vector of each points in a scene.

Combining the two metrics with a proportion $\delta$, we have the final distance metric between points that can be used in clustering:
\begin{equation}
    D(x,y)=\delta \cdot \frac{d_{g}(x,y)}{{\rm average}(d_{g})}+(1-\delta) \cdot \frac{d_{a}(x,y)}{{\rm average}(d_{a})}
\end{equation}

\subsubsection{Spectral Clustering}

In this section, we introduce the subsequent workflow of spectral clustering performed with the proposed distance metric.

Firstly, to reduce calculation complexity, we simplify the 3D mesh with $Quadric\ Edge\ Collapse\ Decimation$ \cite{garland1997surface} \cite{muntoni2021pymeshlab}.
Then, with the proposed distance metric, we calculate a distance matrix (denoted as $D$) which is then smoothed by an exponential kernel and normalized.
The resulting matrix is the affinity matrix of the scene (denoted as $\widetilde{\mathcal{Z}}$):
\begin{equation}
    \begin{aligned}
        D                       & = (d_{ij})_{N\times N},\ d_{ij} = D(i,j)                             \\
        \mathcal{Z}             & = (z_{ij})_{N\times N},\ z_{ij} = e^{-d_{ij} / 2 \sigma^{2}}         \\
        W                       & = {\rm diag}\{ \sum_{i}z_{i1}, \sum_{i}z_{i2}, ... \sum_{i}z_{iN} \} \\
        \widetilde{\mathcal{Z}} & = W^{-\frac{1}{2}}\mathcal{Z}W^{-\frac{1}{2}}                        \\
    \end{aligned}
\end{equation}
Here $N$ is the number of points in the scene, and $\sigma$ is a parameter for smoothing.
In our experiments, we take $\sigma=\frac{1}{N^{2}} \sum_{i, j} D(i, j)$.

Then, we compute $k$ eigenvectors of $\widetilde{Z}$ that have largest eigenvalues, organize them as a matrix $V$ and normalize each row of $V$ to obtain $\widetilde{V}$:
\begin{equation}
    \begin{aligned}
        V             & = [v_1, v_2, ..., v_k] = [a_1, a_2, ..., a_n]^T           \\
        \widetilde{V} & = [\widetilde{a_1},\widetilde{a_2},...,\widetilde{a_n}]^T
    \end{aligned}
\end{equation}
Each row of $\widetilde{V}$ is regarded as the embeddings of the points.

By taking the embeddings of points with ground-truth labels as initial cluster centers, we perform K-means clustering on the embeddings and obtain the spectrum distances between each point and its cluster center.
We then map these distances to the point cloud before downsampling.
The complete pipeline of Spectral Clustering is also demonstrated in Algorithm. \ref{algorithm:2}.

We call a point is \textit{localized} when its spectrum distance to the cluster center is relatively small, as smaller spectrum distances imply more semantic locality information.
In order to effectively select localized pseudo labels with the spectral statistics we harvest, we propose a double mixture probability model similar to Section \ref{sec: unc analysis}.
For the two individual components we choose beta distribution instead, since the normalized spectrum distances are distributed more evenly in $[0,1]$, and the two components represent the localized and unlocalized parts, respectively.
The probablity density of the spectrum distance being $x$ is then given by:
\begin{equation}
    \begin{aligned}
        P(x|\theta_1, \theta_2) & = \omega_1\cdot P(x|\theta_1) + \omega_2\cdot P(x|\theta_2)                    \\
        P(x|\theta)             & = \frac{\Gamma(a+b)}{\Gamma(a)\Gamma(b)}x^{a-1}(1-x)^{b-1},\ \theta = \{a, b\}
    \end{aligned}
\end{equation}
Here, $\omega_1$ is the probability that a sample is drawn from the localized component and $\omega_2$ is the probability that a sample is drawn from the unlocalized component.
$\theta_1$ and $\theta_2$ are the parameter sets of the two components.
Each parameter set $\theta$ consists of two independent parameters $a$ and $b$ controlling the shape of the distribution.

We empirically choose the beta distribution to model the two components of the mixtures, as we observe non-zero skewness in the distribution of spectrum distances.
Similar to Section \ref{sec: unc analysis}, we fit a double mixture model for each category.
Then, for each generated label, we calculate the probabilities that the label is drawn from the localized and unlocalized components, and regard it as a pseudo label if the former is higher than the latter.

The detailed derivation of fitting the double mixture model is provided in Sec. \ref{sec: double model fitting}.

\begin{algorithm}
    \caption{Pipeline for Spectral Analysis}\label{algorithm:2}
    \KwIn{
        \begin{itemize}
            \item [1] $\mathbb{X}$: a large-scale 3D mesh dataset
            \item [2] $\mathcal{T}(\cdot)$: $Quadric\ Edge\ Collapse\ Decimation$
            \item [3] $\mathcal{K}(\cdot)$: K-means clustering
            \item [4] $\mathcal{M}(\cdot)$: Data mapping by nearest neighbor
        \end{itemize}
    }
    \KwOut{$\{d_i\}$: the distance between each point and its cluster center in spectral space}
    \ForEach{$X \in \mathbb{X}$}{
    \tcc{3D mesh simplification}

    $\ \ X' = \mathcal{T}(X)$

    \tcc{Distances between points}

    $\ \ D(x,y)=\delta \cdot \frac{d_{g}(x,y)}{{\rm average}(d_{g})}+(1-\delta) \cdot \frac{d_{a}(x,y)}{{\rm average}(d_{a})}$

    \tcc{Affinity matrix}

    $\ \ \mathcal{Z} = (z_{ij})_{N'\times N'},\ z_{ij} = e^{-D(i,j) / 2 \sigma^{2}}$

    $\ \ W = {\rm diag}\{ \sum_{i}z_{i1}, \sum_{i}z_{i2}, ... \sum_{i}z_{iN'} \}$

    $\ \ \widetilde{Z}=W^{-\frac{1}{2}} \mathcal{Z} W^{-\frac{1}{2}}$

    \tcc{Obtain k eigenvectors with largest eigenvalues}

    $\ \ V=[{v_1},{v_2},...,{v_k}]=[a_1, a_2, ..., a_{N'}]^T$

    \tcc{Normalize each row of the eigenvector matrix}

    $\ \ \widetilde{V}=[\widetilde{{a}}_{1},\widetilde{{a}}_{2},...,\widetilde{{a}}_{N'}]^T$



    \tcc{K-Means clustering}

    $\ \ {d_1', d_2', ..., d_{N'}'} = \mathcal{K}(\{\widetilde{a_1}, \widetilde{a_2}, ..., \widetilde{a_{N'}}\})$

    \tcc{Distances mapping to original 3D mesh}

    $\ \ {d_1, d_2, ..., d_N} = \mathcal{M}(\{d_1', d_2', ..., d_{N'}'\})$
    }
\end{algorithm}



\subsection{Spectrum-Uncertainty Joint Modeling}
\label{sec:joint modeling}

In order to fully exploit the potential of large-scale datasets, we hope to simultaneously make use of both the uncertainty and spectral statistics to harvest pseudo labels with higher reliability.
Therefore, we investigate the distribution of the joint feature of uncertainty and spectrum distances, and propose a joint double mixture model to fit this bivariate distribution ($x$ and $y$ denotes the uncertainty value and the spectrum distance respectively):
\begin{equation}
    \begin{aligned}
        P(x,y|\theta_1, \theta_2) & = \omega_1\cdot P(x, y|\theta_1) + \omega_2\cdot P(x, y|\theta_2)                                            \\
        P(x, y|\theta)            & = \frac{\Gamma(a+b)x^{a-1}(1-x)^{b-1}}{\Gamma(a)\Gamma(b)}\cdot \frac{d^{c}x^{c-1}{\rm e}^{-dx}}{\Gamma(c)}, \\
                                  & \quad\quad\theta = \{a, b, c, d\}
    \end{aligned}
\end{equation}
$\theta_1$ and $\theta_2$ are the parameter sets of the two components.
Each parameter set $\theta$ consists of four independent parameters ($a$, $b$, $c$ and $d$).

Similar to the other two proposed models, this model is also the weighted sum of a reliable component and an unreliable component, and the weights are denoted by $\omega_1, \omega_2$ respectively.
However, this model takes both dropout-based uncertainty and spectrum distance as input and generates 2-dimension distribution probabilities.
With spectrum-uncertainty joint modeling, we take generated labels with higher probabilities to be drawn from the reliable parts as the pseudo labels.
We hope that pseudo labels can be harvested with higher precision by jointly fitting the model statistics of two dimensions.



\subsection{Double Mixture Model Fitting}
\label{sec: double model fitting}

In this section, we introduce the math derivations of fitting double mixture models.

Since the three mixture models we proposed have similar forms, we hereby make the following abstractions:
\begin{itemize}
    \item $\mathbf{x}_i$ is the input (uncertainty, spectrum distance or a pair of the first two) of the i-th point of the scene
    \item $\theta_1, \theta_2$ are the parameter sets of the two components of the mixture model
    \item $p(\cdot|\theta_1, \theta_2)$ is the probability density function given the two parameter sets $\theta_1, \theta_2$
\end{itemize}

Then, our goal is to maximize the log likelihood objective:

\begin{equation}
    \begin{aligned}
        \mathcal{L}(\theta_1, \theta_2) = \sum_{i}\log{p(\mathbf{x} _i|\theta_1, \theta_2})
    \end{aligned}
\end{equation}

Since we use the mixture model, it is difficult to determine which one of the components the input sample belongs to.
So similar to \cite{zhao2020pointly}, we introduce latent variables $\mathbf{Z}=\{\mathbf{z_{i}}\}$.
If $\mathbf{z_{i}}=1$, the sample $\mathbf{x_i}$ is drawn from the reliable (certain, localized) part, otherwise if $\mathbf{z_i}=2$, the sample $\mathbf{x_i}$ is drawn from the unreliable (uncertain, unlocalized) part.
Then the target to optimize is transformed into
\begin{equation}
    \begin{aligned}
        \quad & \mathcal{L}(\theta_1, \theta_2)                                                                                                             \\
        =     & \sum_{i}\log\sum_{\mathbf{z}_i\in \{1,2\}}{\left(p(\mathbf{x} _i|\mathbf{z}_i,\theta_1, \theta_2)p(\mathbf{z}_i|\theta_1, \theta_2)\right)}
    \end{aligned}
\end{equation}

The probability of a sample taken from either of the two parts is independent of the sample itself.
Therefore, we treat these two probabilities as parameters to optimize and refer to them as $\omega_1$ and $\omega_2$.
More specifically, $\omega_1$ denotes the probability that a sample is drawn from the reliable part, and $\omega_2$ denotes the probability that a sample is drawn from the unreliable part.

That being said, we rewrite our target to
\begin{equation}
    \begin{aligned}
        \quad & \mathcal{L}(\theta_1, \theta_2, \omega_1, \omega_2)                                                                                   \\
        =     & \ \sum_{i}\log\sum_{\mathbf{z}_i\in \{1,2\}}{\left(p(\mathbf{x} _i|\mathbf{z}_i,\theta_1, \theta_2)\cdot\omega_{\mathbf{z}_i}\right)}
    \end{aligned}
\end{equation}
and adopt an iterative method to maximize the target by maximizing the difference of $\mathcal{L}(\theta_1, \theta_2, \omega_1, \omega_2)$ in each optimization step:
\begin{equation}
    \begin{aligned}
        \quad & \mathcal{L}(\theta_1, \theta_2, \omega_1, \omega_2) - \mathcal{L}(\theta_1^n, \theta_2^n, \omega_1^n, \omega_2^n)                                                                                                                                                      \\
        =     & \sum_{i}\log\sum_{\mathbf{z}_i\in\{1,2\}}{\frac{p(\mathbf{x} _i|\mathbf{z}_i,\theta_1,\theta_2)\omega_{\mathbf{z}_i}}{p(\mathbf{x}_i|\theta_1^n, \theta_2^n)}}                                                                                                         \\
        =     & \sum_{i}\log\sum_{\mathbf{z}_i\in\{1,2\}}{\frac{p(\mathbf{x} _i|\mathbf{z}_i,\theta_1,\theta_2)\omega_{\mathbf{z}_i} p(\mathbf{z}_i|\mathbf{x}_i, \theta_1^n, \theta_2^n)}{p(\mathbf{x}_i|\theta_1^n, \theta_2^n)p(\mathbf{z}_i|\mathbf{x}_i,\theta_1^n, \theta_2^n)}} \\
    \end{aligned}
\end{equation}
In the above equation, $n$ denotes the number of steps.

Based on the fact that $\sum_{\mathbf{z}_i}p(\mathbf{z}_i|\mathbf{x}_i,\theta_1^n, \theta_2^n)=1$, we use Jensen's inequality and introduce a new notation $\mathcal{L}_n$ to get:
\begin{equation}
    \begin{aligned}
        \quad & \mathcal{L}(\theta_1, \theta_2, \omega_1, \omega_2) - \mathcal{L}(\theta_1^n, \theta_2^n, \omega_1^n, \omega_2^n)                                                                                                                                                 \\
        \ge   & \ \mathcal{L}_{n}(\theta_1, \theta_2, \omega_1, \omega_2)                                                                                                                                                                                                         \\
        =     & \ \sum_{i}\sum_{\mathbf{z}_i}p(\mathbf{z}_i|\mathbf{x} _i, \theta_1^n, \theta_2^n)\log{\frac{p(\mathbf{x} _i|\mathbf{z}_i,\theta_1,\theta_2)\omega_{\mathbf{z}_i}}{p(\mathbf{z}_i|\mathbf{x} _i, \theta_1^n, \theta_2^n)p(\mathbf{x} _i|\theta_1^n, \theta_2^n)}} \\
        =     & \ \sum_{i}E_{\mathbf{z}_i|\mathbf{x} _i, \theta_1^n, \theta_2^n}\left(\log{\frac{p(\mathbf{x} _i|\mathbf{z}_i,\theta_1,\theta_2)\omega_{\mathbf{z}_i}}{p(\mathbf{z}_i|\mathbf{x} _i, \theta_1^n, \theta_2^n)p(\mathbf{x} _i|\theta_1^n, \theta_2^n)}}\right)      \\
    \end{aligned}
\end{equation}

Thus, by increasing $\mathcal{L}_{n}(\theta_1, \theta_2, \omega_1, \omega_2)$ the initial target would also increase.
Leaving out irrelevant terms in $\mathcal{L}_{n}(\theta_1, \theta_2, \omega_1, \omega_2)$, we have

\begin{equation}
    \begin{aligned}
        \quad & {\mathcal{L}_{n}}'(\theta_1, \theta_2, \omega_1, \omega_2)                                                                                                              \\
        =     & \sum_{i}\sum_{\mathbf{z}_i}p(\mathbf{z}_i|\mathbf{x} _i, \theta_1^n, \theta_2^n)\log{\left(p(\mathbf{x} _i|\mathbf{z}_i,\theta_1,\theta_2)\omega_{\mathbf{z}_i}\right)} \\
        =     & \sum_{i}\sum_{\mathbf{z}_i}p(\mathbf{z}_i|\mathbf{x} _i, \theta_1^n, \theta_2^n)\log{\left(p(\mathbf{x} _i|\mathbf{z}_i,\theta_1,\theta_2)\right)}                      \\
        \quad & + \sum_{i}\sum_{\mathbf{z}_i}p(\mathbf{z}_i|\mathbf{x} _i, \theta_1^n, \theta_2^n)\log{\omega_{\mathbf{z}_i}}                                                           \\
    \end{aligned}
\end{equation}
To simplify the expressions, we introduce a set of new notations:
\begin{equation}
    \begin{aligned}
        \mathcal{J}_{n,1}(\theta_1, \theta_2) & = \sum_{i}\sum_{\mathbf{z}_i}p(\mathbf{z}_i|\mathbf{x} _i, \theta_1^n, \theta_2^n)\log{\left(p(\mathbf{x} _i|\mathbf{z}_i,\theta_1,\theta_2)\right)} \\
        \mathcal{J}_{n,2}(\omega_1, \omega_2) & = \sum_{i}\sum_{\mathbf{z}_i}p(\mathbf{z}_i|\mathbf{x} _i, \theta_1^n, \theta_2^n)\log{\omega_{\mathbf{z}_i}}                                        \\
    \end{aligned}
\end{equation}

Then we update the parameters iteratively:
\begin{equation}
    \begin{aligned}
        \quad & \theta_1^{n+1}, \theta_2^{n+1}, \omega_1^{n+1}, \omega_2^{n+1}                                                                                              \\
        =     & \ \underset{\theta_1, \theta_2, \omega_1, \omega_2}{\rm argmax}\left\{\mathcal{L}_{n}(\theta_1, \theta_2, \omega_1, \omega_2)\right\}                       \\    =&\ \underset{\theta_1, \theta_2, \omega_1, \omega_2}{\rm argmax}\left\{\mathcal{L}_{n}'(\theta_1, \theta_2, \omega_1, \omega_2)\right\} \\
        =     & \ \underset{\theta_1, \theta_2, \omega_1, \omega_2}{\rm argmax}\left\{\mathcal{J}_{n,1}(\theta_1, \theta_2) + \mathcal{J}_{n,2}(\omega_1, \omega_2)\right\} \\
    \end{aligned}
\end{equation}

Since $\theta_1, \theta_2$ are independent of $\omega_1, \omega_2$, we update the two sets of parameters independently:
\begin{equation}
    \begin{aligned}
        \theta_1^{n+1}, \theta_2^{n+1} & = \underset{\theta_1, \theta_2}{\rm argmax}\left\{\mathcal{J}_{n,1}(\theta_1, \theta_2))\right\} \\
        \omega_1^{n+1}, \omega_2^{n+1} & = \underset{\omega_1, \omega_2}{\rm argmax}\left\{\mathcal{J}_{n,2}(\omega_1, \omega_2))\right\} \\
    \end{aligned}
\end{equation}

We adopt an EM algorithm to monotonically increase $\mathcal{J}_{n,1}$ and $\mathcal{J}_{n,2}$.
In each iteration, we first calculate $\mathcal{J}_{n,1}, \mathcal{J}_{n,2}$ with inputs and current parameters, and then update the parameters by assigning the gradients of $\mathcal{J}_{n_1}, \mathcal{J}_{n_2}$ to zero.
After a fixed number of iterations, the fitting process converges.

\subsection{Fine-tuning with Pseudo Labels}
\label{sec:pslgen}

With the pseudo labels we harvest, we fine-tune the pre-trained Viewpoint Bottleneck framework again.
For each point in the scene, we report the category with the maximum logit as the prediction for semantic segmentation.

\section{Experiments}

\subsection{Datasets}
\noindent \textbf{Dataset for SSRL}\quad
ScanNetV2 \cite{hou2021exploring} is a large-scale RGB-D video dataset with 3D reconstructions of indoor scenes with over 1500 scans reconstructed from around 2.5 million views.
The dataset also provides point-level semantic labels for the scenes, which are defined in a protocol of 20 categories.
In our study, we conduct the experiments of self-supervised representation pre-training on the official training split of ScanNetV2 \cite{dai2017scannet} that contains 1201 scans. Our method is evaluated on the official validation split (312 scans) of the dataset.We also provide quantitative results on the held-out test set, which is reported by the official online server.

Stanford Large-Scale 3D Indoor Spaces (S3DIS) \cite{7780539} dataset containes 3D scans of six indoor areas captured from 3 different buildings. Each area is scanned with RGBD sensors and is represented as point clouds with xyz coordinate and RGB value and annotated with 13 object categories. The evaluation protocol on Area 5 as held-out is adopted. We use 5cm voxel for the experiments.

Semantic3D \cite{hackel2017isprs} is a large labelled 3D point cloud dataset of natural scenes, which collected by static terrestrial laser scanners. It contains 8 semantic classes and covers diverse urban scenes. In the commonly used benchmark, there are 9 point clouds in the training set and 6 point clouds in the validation set.

\noindent \textbf{Dataset for data-efficient fine-tuning}\ \
For fair comparison, we use the data-efficient settings of ScannetV2 provided by \cite{hou2021exploring} for fine-tuning.
In these settings, only 20/50/100/200 points of a scene in the training splits have labels.
Our method leverages the sparse ground-truth annotations to generate reliable pseudo labels.
The pseudo labels then serve as supervision during fine-tuning, and other unlabeled points are ignored.
For weakly supervised settings of S3DIS and Semantic3D datasets, we evaluate on a subset of points (20/50/100/200 points) uniformly sampled within each points cloud.
On these validation sets, the intersection over union (IoU) is taken as the evaluation metrics


\subsection{Implementation Details}

To prepare the scene meshes for further exploitation, we sample a combination of the following transformations to apply to the point clouds:
\begin{itemize}
    \item Random rotation. We rotate the point cloud along z-axis by a random angle uniformly sampled from $[0, 2\pi]$.
    \item Random mirroring. For each axis, we apply mirroring transformation with a probability of 0.5.
    \item Random chromatic jitter. For each channel, we apply a noise sampled from $\mathcal{N}(0, 255 \times 0.05)$.
\end{itemize}

Then, the point clouds are downsampled with farthest point sampling (FPS) before being fed to subsequent neural networks.
By always selecting the point that is farthest to the already chosen ones, FPS ensures that only redundant points are discarded and the downsampled point cloud is an abstraction of the original point cloud.
The target size of downsampled point cloud is 1024 points due to the limitation of GPU memory size used in subsequent neural networks.

We choose Sparse ConvUNet as the backbone of the neural network and implement it with MinkowskiNetwork \cite{choy20194d}.
Sparse ConvNet has relatively small GPU memory footprints, which enables deeper network and well suits the scenarios where large amounts of points need to be processed simultaneously.
The network takes both 3D coordinates and point-wise RGB values as the inputs.
The voxel size for Sparse ConvUNet is set to 2.0 cm for the ScannetV2 dataset and 5.0cm for the S3DIS dataset, since the distances between points in these datasets are within 20.0 cm. For the Semantic3D dataset, we set the voxel size to 50.0 cm due to the larger distances between points (up to 2.0 m).
The feature dimension is set to 256/512/1024 to investigate the robustness of Viewpoint Bottleneck.

For SSRL with Viewpoint Bottleneck, experiments are conducted with a batch size of 2 for 20000 iterations on two GeForce RTX 3090 GPUs.
An SGD optimizer with a momentum of 0.99 is adopted.
The learning rate is decayed polynomially, with the initial value set to 0.1.

For the first and second fine-tuning processes, experiments are conducted with a batch size of 12 for 30000 iterations on two GeForce RTX 3090 GPUs with the same optimizer and scheduler settings.

For uncertainty modeling, we obtain the uncertainty of the labels by running with active dropout layers 10 times with a dropout rate of 0.5.
For spectrum modeling, we set the downsample target to 8000, geodesic-to-angular distance proportion to 3:2, and the dimension of the feature space to 50 to optimize the performance.
We adopt the implementation in scikit-learn library \cite{sklearn_api} (i.e. \textit{sklearn.cluster.KMeans}) for the K-means clustering.

When fitting the mixture models, we adopt the EM algorithm with 50 iterations which is enough for the algorithm to converge.

\subsection{Results and Analysis}

\begin{table}[htbp]
    \centering
    \caption{3D Scene Semantic Segmentation Results on ScanNet Dataset}
    \scalebox{0.6}{
        \begin{tabular}{cccc}
            \toprule
            \multirow{2}[2]{*}{Setting}            & \multirow{2}[2]{*}{Method}               & \multicolumn{2}{c}{Split}                 \\
            \cmidrule{3-4}                         &                                          & val.                      & test          \\
            \midrule
            \multirow{5}[2]{*}{Fully}              & PointNet++\cite{NIPS2017_d8bf84be}       & -                         & 33.9          \\
                                                   & PointCNN\cite{Li2018PointCNNCO}          & -                         & 45.8          \\
                                                   & KPConv\cite{9010002}                     & -                         & 68.4          \\
                                                   & MinkowskiNet\cite{choy20194d}            & -                         & 73.6          \\
                                                   & BPNet\cite{hu-2021-bidirectional}        & -                         & 74.9          \\
            \midrule
            \multirow{2}[2]{*}{subs.}              & WyPR\cite{9577806}                       & 31.1                      & 24.0          \\
                                                   & MPRM\cite{9157503}                       & 43.2                      & 41.1          \\
            \midrule
            10{}\%                                 & WS3\cite{zhang2021weakly}                & -                         & 52.0          \\
            \midrule
            \multirow{3}[2]{*}{1{}\%}              & WS3\cite{zhang2021weakly}                & -                         & 51.1          \\
                                                   & HybridCR\cite{Li_2022_CVPR}              & 56.9                      & 56.8          \\
                                                   & PSD\cite{9710706}                        & -                         & 54.7          \\
            \midrule
            \multirow{4}[2]{*}{200 pts (0.1{}\%)}  & SQN\cite{9710706}                        & -                         & 59.8          \\
                                                   & CSC\cite{hou2021exploring}               & 68.2                      & 66.5          \\
                                                   & PointContrast\cite{xie2020pointcontrast} & -                         & 65.3          \\
                                                   & \textbf{VIB(ours)}                       & \textbf{68.5}             & \textbf{66.9} \\
                                                   & \textbf{VIBUS(ours)}                     & \textbf{69.6}             & \textbf{69.1} \\
            \midrule
            \multirow{4}[2]{*}{100 pts (0.05{}\%)} & SQN\cite{9710706}                        & -                         & 57.6          \\
                                                   & CSC\cite{hou2021exploring}               & 65.9                      & 64.4          \\
                                                   & PointContrast\cite{xie2020pointcontrast} & -                         & 63.6          \\
                                                   & \textbf{VIB(ours)}                       & \textbf{66.8}             & \textbf{65.0} \\
                                                   & \textbf{VIBUS(ours)}                     & \textbf{68.9}             & \textbf{68.4} \\
            \midrule
            \multirow{4}[2]{*}{50 pts (0.03{}\%)}  & SQN\cite{9710706}                        & -                         & 54.2          \\
                                                   & CSC\cite{hou2021exploring}               & 60.5                      & 61.2          \\
                                                   & PointContrast\cite{xie2020pointcontrast} & -                         & 61.4          \\
                                                   & \textbf{VIB(ours)}                       & \textbf{63.6}             & \textbf{62.3} \\
                                                   & \textbf{VIBUS(ours)}                     & \textbf{65.6}             & \textbf{65.1} \\
            \midrule
            \multirow{5}[2]{*}{20 pts (0.01{}\%)}  & SQN\cite{9710706}                        & -                         & 48.6          \\
                                                   & CSC\cite{hou2021exploring}               & 55.5                      & 53.1          \\
                                                   & PointContrast\cite{xie2020pointcontrast} & -                         & 55.0          \\
                                                   & DAT\cite{wu2022dual}                     & 58.6                      & -             \\
                                                   & \textbf{VIB(ours)}                       & \textbf{57.0}             & \textbf{54.8} \\
                                                   & \textbf{VIBUS(ours)}                     & \textbf{61.0}             & \textbf{58.6} \\
            \bottomrule
        \end{tabular}
    }
    \label{tbl:scannet_ss}%
\end{table}

\maketable{scannet_ss_lr}{0.7}{ht}{}
{\centering 3D Scene Semantic Segmentation Result on ScanNet Validation Set (Trained with Limited Reconstructions, mIOU: \%)}
{
    \begin{tabular}{lcccc}
        \toprule
        {Points}                      & {20\%}        & {10\%}        & {5\%}         & {1\%}         \\
        \midrule
        {Scratch}                     & {48.1}        & {42.7}        & {31.9}        & {9.9}         \\
        {CSC \cite{hou2021exploring}} & {50.6(+2.5)}  & {44.9(+2.2)}  & {36.6(+4.4)}  & {13.2(+3.3)}  \\
        \midrule
        {VIB(ours)}                   & {64.8(+16.7)} & {60.5(+17.8)} & {47.4(+10.8)} & {28.6(+18.7)} \\
        \bottomrule
    \end{tabular}
}

\maketable{scannet_is}{0.7}{ht}{}
{{3D Scene Instance Segmentation Results on ScanNet Validation Set }}
{
    \begin{tabular}{lcccc}
        \toprule
        {Setting}          & {Method}                                         & {mAP@50\%}    \\
        \midrule
        {Fully}            & {MTML\cite{9008793}}                             & {40.2}        \\
        {Fully}            & {3D-MPA\cite{9156487}}                           & {59.1}        \\
        {Fully}            & {OccuSeg\cite{9157103}}                          & {60.7}        \\
        {Fully}            & {SSTNet\cite{Liang_2021_ICCV}}                   & {64.3}        \\
        \midrule
        {Boxes}            & {Box2Mask\cite{chibane2021box2mask}}             & {59.7}        \\
        {20 pts (0.01\%)}  & {CSC\cite{hou2021exploring}}                     & {26.3}        \\
        {20 pts (0.01\%)}  & \textbf{VIB(ours)}                               & \textbf{35.2} \\
        {50 pts (0.03\%)}  & {CSC\cite{hou2021exploring}}                     & {32.6}        \\
        {50 pts (0.03\%)}  & \textbf{VIB(ours)}                               & \textbf{39.7} \\
        {100 pts (0.05\%)} & {CSC\cite{hou2021exploring}}                     & {39.3}        \\
        {100 pts (0.05\%)} & \textbf{VIB(ours)}                               & \textbf{43.3} \\
        {200 pts (0.1\%)}  & {PointContrast\cite{xie2020pointcontrast}} & {44.5}        \\
        {200 pts (0.1\%)}  & {CSC\cite{hou2021exploring}}                     & {48.9}        \\
        {200 pts (0.1\%)}  & \textbf{VIB(ours)}                               & \textbf{45.1} \\
        \bottomrule
    \end{tabular}
}

\textbf{Results on ScanNet V2 Dataset}
To show the effectiveness of our proposed method, we evaluate state-of-the-art methods and ours on the ScanNetV2 dataset and report the results in Table. \ref{tbl:scannet_ss}.
Note that, our methods are compared with three groups of settings:
(1) fully supervised methods trained with 100\% point-level annotations (denoted by "Fully")
(2) weakly-supervised methods trained with scene/subcloud-level annotations (denoted by "subs.")
(3) weakly-supervised methods trained with limited point-level annotations.

In the third setting that has the least supervision with up to 200 annotations (0.1\% of the total set of annotations) available per scene, our methods achieve the highest performances on the ScanNet validation set among weakly-supervised methods on the task of semantic segmentation.
With 20/50/100 initial annotations, VIBUS achieves large margins over other methods in this setting, especially in the setting of 20 initial annotations (+6.0 mIoU compared with the CSC\cite{hou2021exploring}).
With 200 initial annotations, our method also outperforms other methods in this setting but with smaller margins due to the saturation of performances.

We also report the performances of VIBUS and some other baselines on the ScanNet test set in Table. \ref{tbl:scannet_ss}, which are provided by the online Scannet Benchmark and ScanNet Data Efficient Benchmark.
Note that, the evaluation on the test split is conducted on the test server as the ground truth labels of the ScanNet test set are not publicly available.
Our method (Spec-Unc Field) ranks 1$^{st}$ on the track with 100 initial annotations.
We also achieve leading performances on all other tracks with 20/50/200 initial annotations, higher than all the baselines.

\begin{figure*}[htbp]
    \centerline{\includegraphics[width=0.9\textwidth]{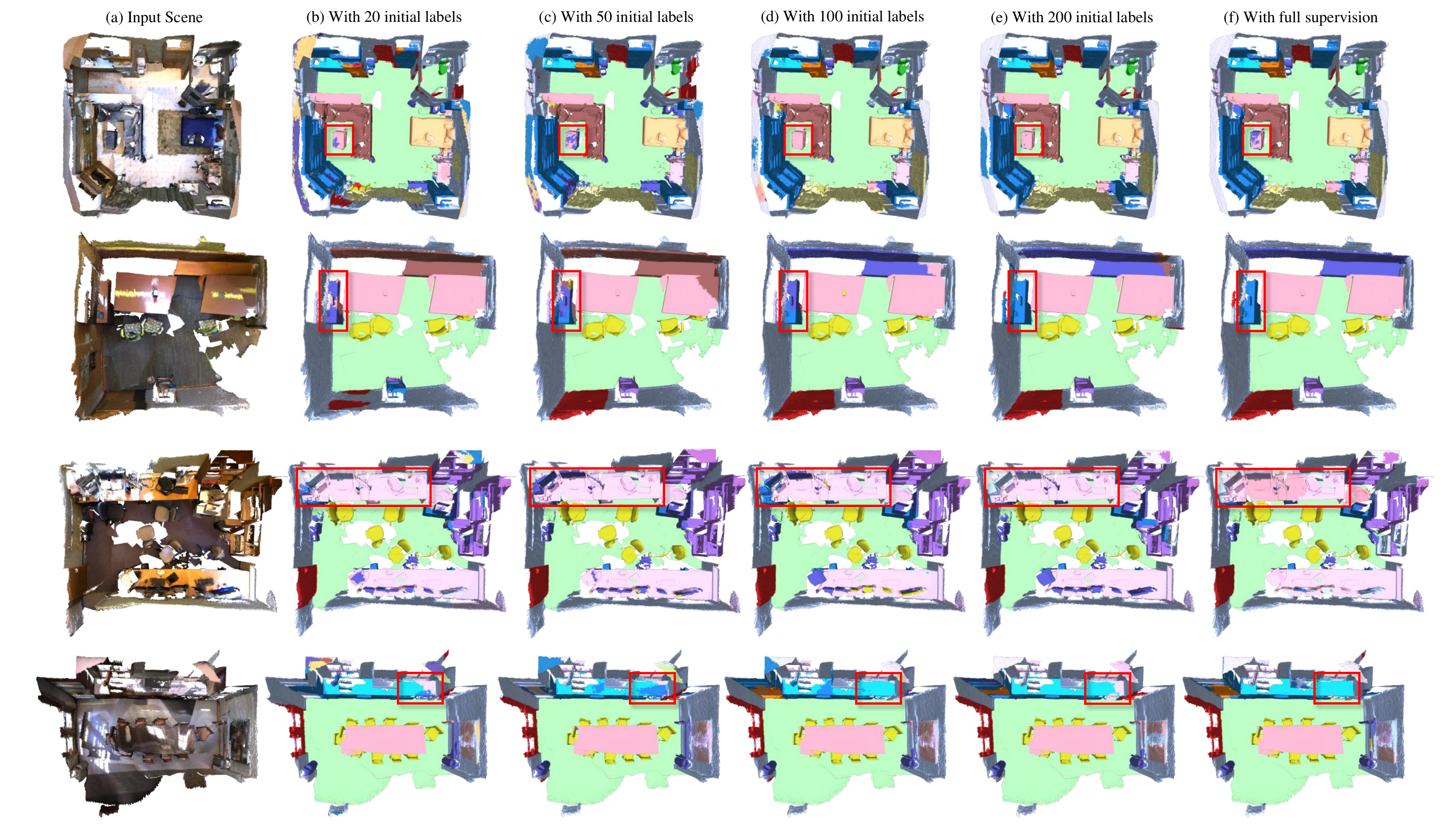}}
    \caption{The semantic segmentation qualitative results of VIBUS on the ScanNet validation set under different levels of supervision.}
    \label{fig:qualitative_visualized}
\end{figure*}

\begin{figure*}[htbp]
    \centerline{\includegraphics[width=1\textwidth]{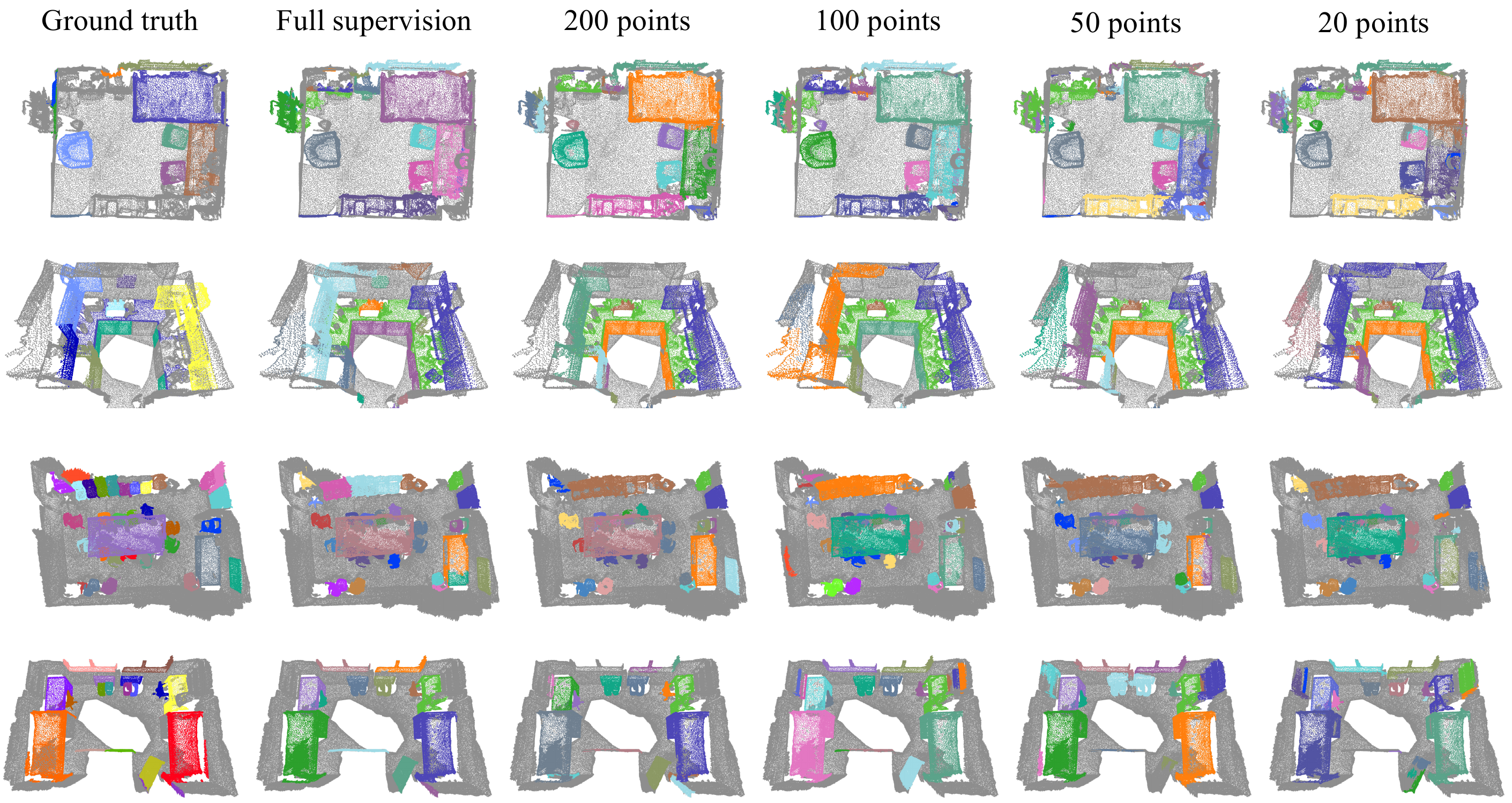}}
    \caption{The qualitative results of Scannet instance segmentation of VIB in the settings with different initial labels.}
    \label{fig:scannet_is}
\end{figure*}

We also show the semantic segmentation results under the limited reconstructions setting in Table \ref{tbl:scannet_ss_lr}.
The performance of VIB achieves significant margins over the baseline CSC\cite{hou2021exploring}.
In the Table. \ref{tbl:scannet_is}, we report the instance segmentation results of different methods on the ScanNetV2 validation set. In weakly-supervised settings, our method achieves better performances compared with CSC\cite{hou2021exploring} but still underperforms fully-supervised methods.
In Fig. \ref{fig:qualitative_visualized}, we show the qualitative results  of VIBUS on the task of semantic segmentation on the ScanNet dataset in the different settings. In Fig. \ref{fig:scannet_is}, we show the qualitative results of VIB on the task of instance segmentation with the different levels of supervision.
These results prove the novelty of our method under extremely sparse supervision.



\textbf{Results on Area-5 of S3DIS}
We evaluate the state-of-the-art methods and our methods on the Area-5 of S3DIS and report the semantic segmentation results in Table. \ref{tbl:s3dis_ss}. 
In the four weakly-supervised settings with which we evaluate our methods, we use up to 200 point annotations per scene which is a small portion (up to 0.02\%) of the whole set of annotations.
In the 100 annotations (0.01\%) setting, our method outperforms MT\cite{Tarvainen2017MeanTA} and Xu et al.\cite{Xu_2020_CVPR} which are evaluated in the 0.2\% setting.
In the 200 annotations (0.02\%) setting, the performance of our method is the best among all other methods which are evaluated in the 0.2\% settings.
In general, our method obtains better performance with fewer annotations, proving the effectiveness and the novelty of our method.

\begin{figure*}[htbp]
    \centerline{\includegraphics[width=1\textwidth]{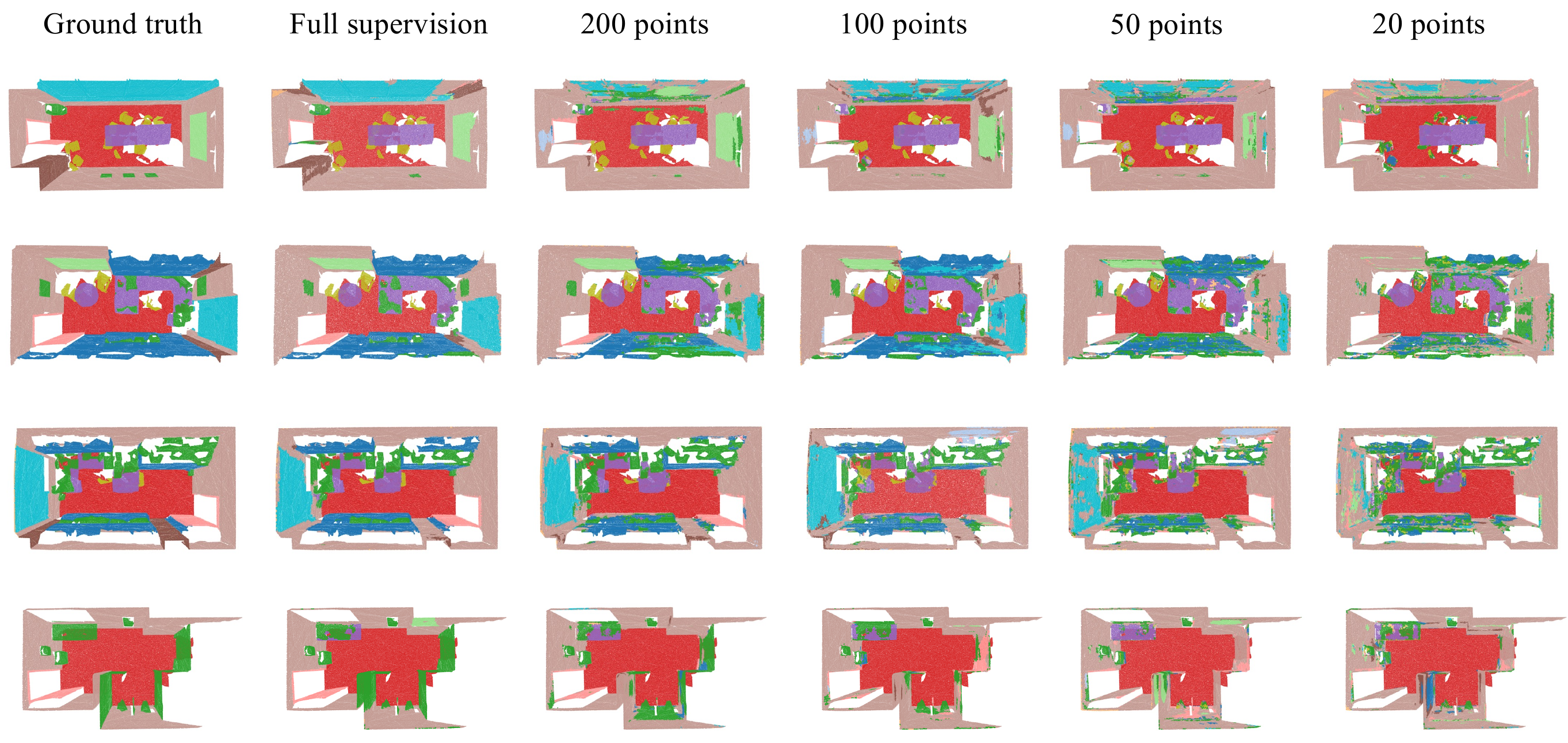}}
    \caption{The qualitative results of S3DIS semantic segmentation of VIB in the settings with different initial labels.}
    \label{fig:s3dis_ss}
\end{figure*}

\begin{figure*}[htbp]
    \centerline{\includegraphics[width=1\textwidth]{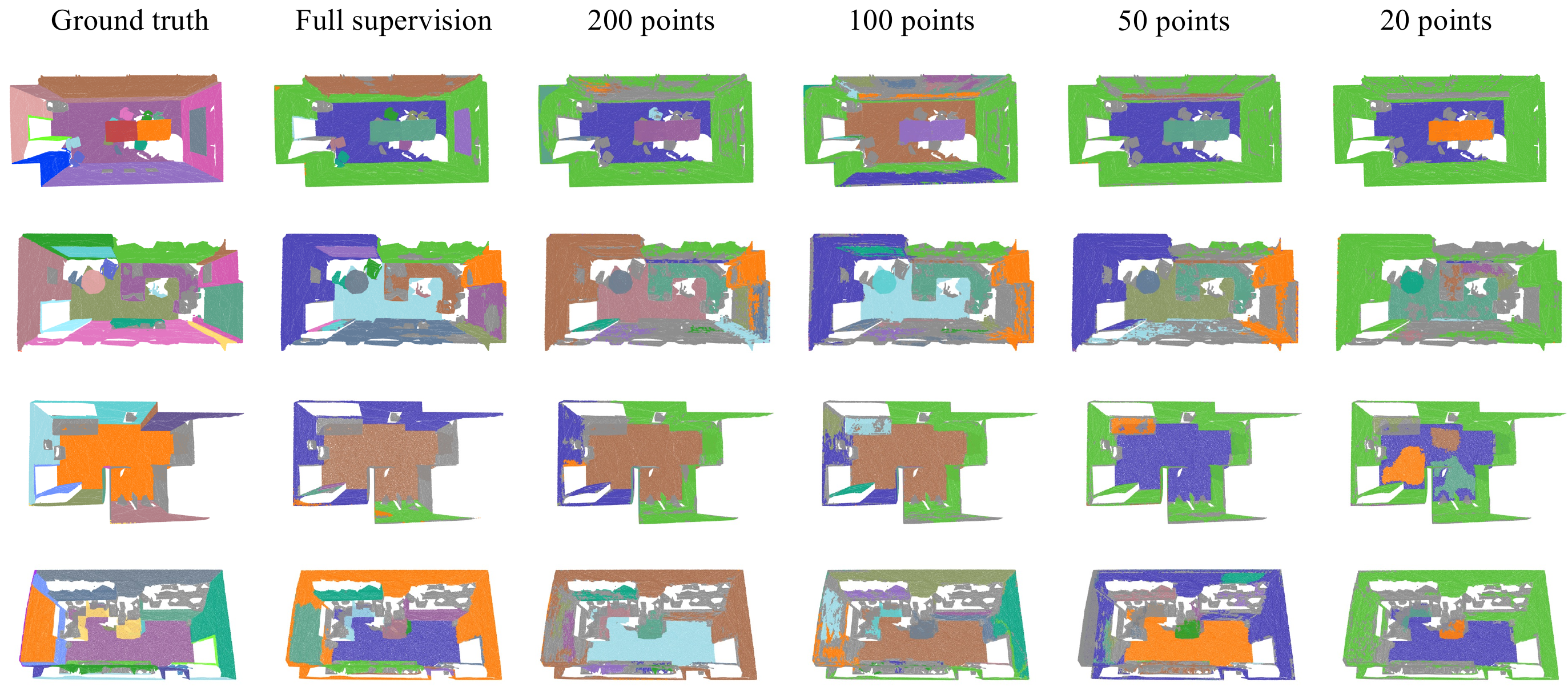}}
    \caption{The qualitative results of S3DIS instance segmentation of VIB in the settings with different initial labels.}
    \label{fig:s3dis_is}
\end{figure*}

We report the performance of our methods on the task of instance segmentation on S3DIS in Table. \ref{tbl:s3dis_is}.
VIB achieves significant margins over the baseline in all weakly supervised settings.
In Fig. \ref{fig:s3dis_ss} and \ref{fig:s3dis_is}, we also show the qualitative results of VIB on the task of semantic segmentation and instance segmentation with different levels of supervision.

\maketable{s3dis_ss}{0.7}{ht}{}
{{3D Scene Semantic Segmentation Results on Area-5 of S3DIS}}
{
    \begin{tabular}{lcccc}
        \toprule
        {Setting}          & {Method}                          & {mIoU} \\
        \midrule
        {Fully}            & {PointCNN\cite{Li2018PointCNNCO}} & {57.3} \\
        {Fully}            & {KPConv\cite{9010002}}            & {67.1} \\
        {Fully}            & {HybridCR\cite{Li_2022_CVPR}}     & {65.8} \\
        {Fully}            & {PSD\cite{9710706}}               & {65.1} \\
        {Fully}            & {RandLA-Net\cite{9156466}}        & {62.4} \\
        \midrule
        {10\%}             & {Xu et al.\cite{Xu_2020_CVPR}}    & {48.0} \\
        {10\%}             & {WS3\cite{zhang2021weakly}}       & {64.0} \\
        {1\%}              & {WS3\cite{zhang2021weakly}}       & {61.8} \\
        {1\%}              & {PSD\cite{9710706}}               & {63.6} \\
        {0.2\%}            & {MT\cite{Tarvainen2017MeanTA}}    & {44.4} \\
        {0.2\%}            & {Xu et al.\cite{Xu_2020_CVPR}}    & {44.5} \\
        {0.1\%}            & {RandLA-Net\cite{9156466}}        & {52.9} \\
        {0.03\%}           & {PSD\cite{9710706}}               & {48.2} \\
        {0.03\%}           & {HybridCR\cite{Li_2022_CVPR}}     & {51.5} \\
        \midrule
        {20 pts (0.002\%)} & {VIB(ours)}                       & {29.1} \\
        {50 pts (0.006\%)} & {VIB(ours)}                       & {38.0} \\
        {100 pts (0.01\%)} & {VIB(ours)}                       & {46.4} \\
        {200 pts (0.02\%)} & {VIB(ours)}                       & {52.0} \\
        \bottomrule
    \end{tabular}
}

\maketable{s3dis_is}{0.7}{ht}{}
{{3D Scene Instance Segmentation Result on Area-5 of S3DIS (Trained with Limited Annotation, mAP@50: \%)}}
{
    \begin{tabular}{lcccc}
        \toprule
        {Points}    & {200}        & {100}        & {50}         & {20}         \\
        \midrule
        {Baseline}  & {20.8}       & {15.0}       & {8.4}        & {5.3}        \\
        {VIB(ours)} & {21.4(+0.6)} & {17.0(+2.0)} & {17.6(+9.2)} & {13.9(+8.6)} \\
        \bottomrule
    \end{tabular}
}

\begin{figure*}[htbp]
    \centerline{\includegraphics[width=1\textwidth]{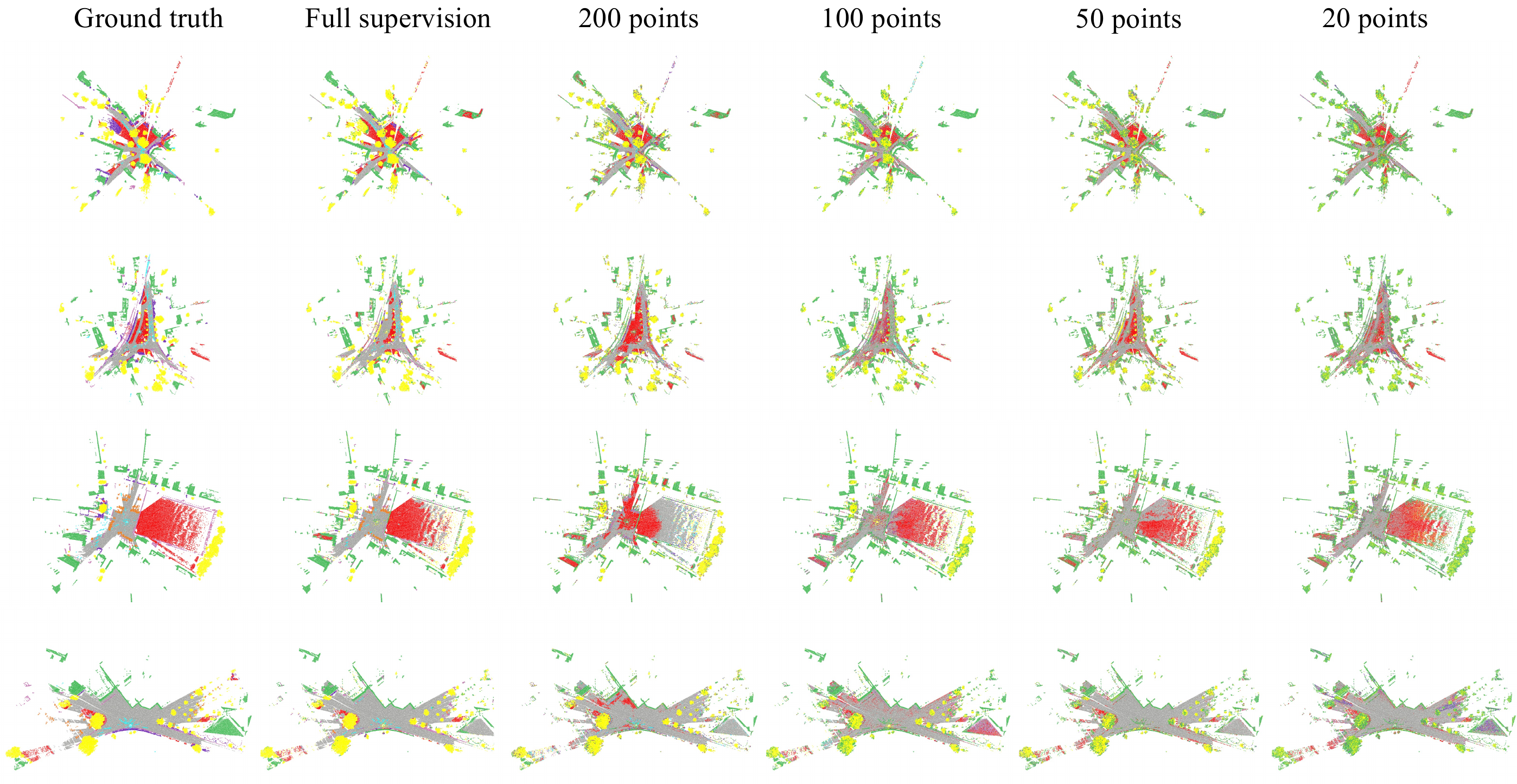}}
    \caption{The qualitative results of Semantic3D semantic segmentation of VIB in the settings with different initial labels.}
    \label{fig:semantic3d_ss}
\end{figure*}

\textbf{Results on Semantic3D (semantic-8)} We evaluate our method on the task of semantic segmentation on Semantic3D (semantic-8) and report the results in Table. \ref{tbl:semantic3D_ss}.
Directly training with the extremely sparse annotations is considered as the baseline.
VIB achieves postive margins over the baselines in different weakly-supervised settings, showing the effectiveness of our method on large-scale outdoor scenes. We show the qualitative results of VIB on the task of semantic segmentation in Fig. \ref{fig:semantic3d_ss}.

\maketable{semantic3D_ss}{0.7}{ht}{}
{{3D Scene Semantic Segmentation Results on Semantic3D (semantic-8)}}
{
    \begin{tabular}{lcccc}
        \toprule
        {Setting}          & {Method}                             & {mIoU} \\
        \midrule
        {Fully}            & {PointNet++\cite{NIPS2017_d8bf84be}} & {63.1} \\
        {Fully}            & {Qiu et al.\cite{9577557}}           & {75.4} \\
        \midrule
        {10\%}             & {WS3\cite{zhang2021weakly}}          & {73.3} \\
        {1\%}              & {WS3\cite{zhang2021weakly}}          & {72.6} \\
        {0.1\%}            & {SQN\cite{9710706}}            & {72.3} \\
        {0.01\%}           & {SQN\cite{9710706}}                  & {58.8} \\
        \midrule
        {20 pts (0.002\%)} & {baseline}                           & {14.4} \\
        {20 pts (0.002\%)} & {VIB(ours)}                          & {22.7} \\
        {50 pts (0.006\%)} & {baseline}                           & {26.0} \\
        {50 pts (0.006\%)} & {VIB(ours)}                          & {33.4} \\
        {100 pts (0.01\%)} & {baseline}                           & {32.6} \\
        {100 pts (0.01\%)} & {VIB(ours)}                          & {34.2} \\
        {200 pts (0.02\%)} & {baseline}                           & {36.0} \\
        {200 pts (0.02\%)} & {VIB(ours)}                          & {39.4} \\
        \bottomrule
    \end{tabular}
}

\maketable{ablation_vib}{0.7}{ht}{}
{Evaluation of Viewpoint Bottleneck$^{\mathrm{{a}}}$ with Different Feature Sizes {on ScanNet}(mIoU: \%)}{
    \begin{tabular}{lcccc}
        \toprule
        Points      & 200           & 100           & 50            & 20            \\
        \midrule
        Ours (256)  & 68.4          & 66.5          & 63.3          & 56.2          \\
        Ours (512)  & \textbf{68.5} & \textbf{66.8} & 63.6          & \textbf{57.0} \\
        Ours (1024) & 68.4          & 66.5          & \textbf{63.7} & 56.3          \\
        \bottomrule
    \end{tabular}
}{
    \begin{tablenotes}
        \scriptsize
        \item[{a}] {Viewpoint Bottleneck is the first stage of VIBUS. The evaluation of Viewpoint Bottleneck is conducted by fine-tuning the first-stage network with limited ground-truth labels only.}
    \end{tablenotes}
}

\maketable{ablation_vibus}{0.7}{ht}{*}
{Evaluation of VIBUS with Different Modeling Strategies (mIoU: \%)}{
    \begin{tabular}{lcccc}
        \toprule
        Points                  & 200           & 100           & 50            & 20            \\
        \midrule
        VIB (512) + Uncertainty & \textbf{69.6} & 68.6          & 64.7          & 60.7          \\
        VIB (512) + Spectrum    & 69.1          & 68.5          & \textbf{65.6} & \textbf{61.0} \\
        VIB (512) + Joint       & 69.4          & \textbf{68.9} & 64.6          & 60.0          \\
        \bottomrule
    \end{tabular}
}

\maketable{eval_pslabels}{0.7}{ht}{}
{Evaluation of Pseudo Labels Harvested by Different Strategies (mIoU: \%)}{
    \begin{tabular}{lcccc}
        \toprule
        Points               & 200           & 100           & 50            & 20            \\
        \midrule
        Spectrum Modeling    & 51.9          & 51.6          & 46.9          & 39.6          \\
        Uncertainty Modeling & \textbf{61.1} & \textbf{59.4} & \textbf{52.8} & \textbf{49.4} \\
        Joint Modeling       & 61.0          & 59.2          & 49.8          & 45.2          \\
        \bottomrule
    \end{tabular}
}

\maketable{ablation_dsrate}{0.7}{ht}{}
{Evaluation of Spectral Clustering with Different Target Vertices Number during Downsampling (mIoU: \%)}{
    \begin{tabular}{cccccccccccccccccccccc}
        \toprule
        Downsample Target & 2,000 & 4,000         & 8,000 & 12,000        & 16,000        & 20,000        \\
        \midrule
        20                & 24.2  & 24.9          & 24.8  & \textbf{25.1} & \textbf{25.1} & \textbf{25.1} \\
        50                & 34.1  & \textbf{34.3} & 34.2  & 34.2          & 34.0          & 33.8          \\
        100               & 41.9  & \textbf{42.7} & 42.6  & 42.4          & 42.4          & 42.3          \\
        200               & 48.1  & 49.3          & 49.6  & \textbf{49.8} & 49.7          & 49.7          \\
        \bottomrule
    \end{tabular}
}

\maketable{ablation_featdim}{0.7}{ht}{}
{Evaluation of Spectral Clustering with Different Embedding Lengths (mIoU: \%)}{
    \begin{tabular}{cccccccccccccccccccccc}
        \toprule
        Embedding Length & 10   & 30   & 50            & 70   & 90            & 110           & 200  \\
        \midrule
        20               & 23.5 & 24.4 & \textbf{25.1} & 24.5 & 24.1          & 23.3          & 20.0 \\
        50               & 31.1 & 33.6 & 34.0          & 33.9 & \textbf{34.2} & \textbf{34.2} & 32.4 \\
        100              & 37.5 & 41.7 & \textbf{42.4} & 41.3 & 40.9          & 40.9          & 40.3 \\
        200              & 43.4 & 48.5 & \textbf{49.7} & 49.1 & 48.4          & 48.1          & 46.6 \\
        \bottomrule
    \end{tabular}
}

\maketable{ablation_gdad}{0.7}{ht}{}
{Evaluation of Spectral Clustering with Different Percentage of Geodesic Distance $\delta$ (mIoU: \%)}{
    \begin{tabular}{cccccccccccccccccccccc}
        \toprule
        Proportion of & \multirow{2}{*}{0.0} & \multirow{2}{*}{0.2} & \multirow{2}{*}{0.4} & \multirow{2}{*}{0.6} & \multirow{2}{*}{0.8} & \multirow{2}{*}{1.0} \\
        Geodesic Distance $\delta$                                                                                                                              \\
        \midrule
        20            & 13.3                 & 24.6                 & \textbf{25.0}        & 24.4                 & 24.2                 & 23.6                 \\
        50            & 15.3                 & 33.6                 & \textbf{34.2}        & 33.6                 & 33.5                 & 32.0                 \\
        100           & 16.0                 & 40.4                 & \textbf{41.7}        & \textbf{41.7}        & 41.5                 & 38.5                 \\
        200           & 16.6                 & 46.2                 & 48.1                 & \textbf{48.5}        & \textbf{48.5}        & 44.8                 \\
        \bottomrule
    \end{tabular}
}

\maketable{time}{0.7}{ht}{}
{Time Consumed in Different Stages}{
    \begin{tabular}{ccccc}
        \toprule
        {pre-training} & {fine-tuning} & {modeling} & {inference} \\
        \midrule
        {3 days}       & {1 day}       & {6 hours}  & {5 minutes} \\
        \bottomrule
    \end{tabular}
}

\subsection{Ablations}
To demonstrate the robustness of our method, we conduct several experiments for ablation.

\textbf{Feature space dimension in for Viewpoint Bottleneck} Table. \ref{tbl:ablation_vib} shows the performances of Viewpoint Bottleneck on the validation set with different feature space dimension.
We argue that Viewpoint Bottleneck is robust to the feature space dimension since little effect is observed on the final mIoUs.
In other experiments, the feature space dimension is set to 512 due to the slight improvements over the others.

\textbf{Modeling strategies} We further explore the effectiveness of VIBUS with different strategies for pseudo label harvesting and report the performances on the Scannet semantic segmentation validation set in Table. \ref{tbl:ablation_vibus}, and the semantic segmentation of per-category performance in Table \ref{tab:class_iou}.
We denote Viewpoint Bottleneck as VIB and denote uncertainty, spectrum, {and} joint modeling strategies as 'Uncertainty', 'Spectrum', and 'Joint'.
In addition, we examine the quality of pseudo labels by directly calculating the mIoU of the pseudo label sets.
The results are reported in Table. \ref{tbl:eval_pslabels}.

Results show that the three strategies produce pseudo labels with different mIoU.
Specifically, uncertainty modeling produces pseudo labels with the highest mIoU, and spectrum modeling produces pseudo labels with the lowest mIoU.
However, fine-tuning with the three sets of pseudo labels achieves comparable final performances.
This phenomenon indicates that the three approaches are all capable of filtering reliable pseudo labels.

\begin{figure*}[htbp]
    \centerline{\includegraphics[width=0.9\textwidth]{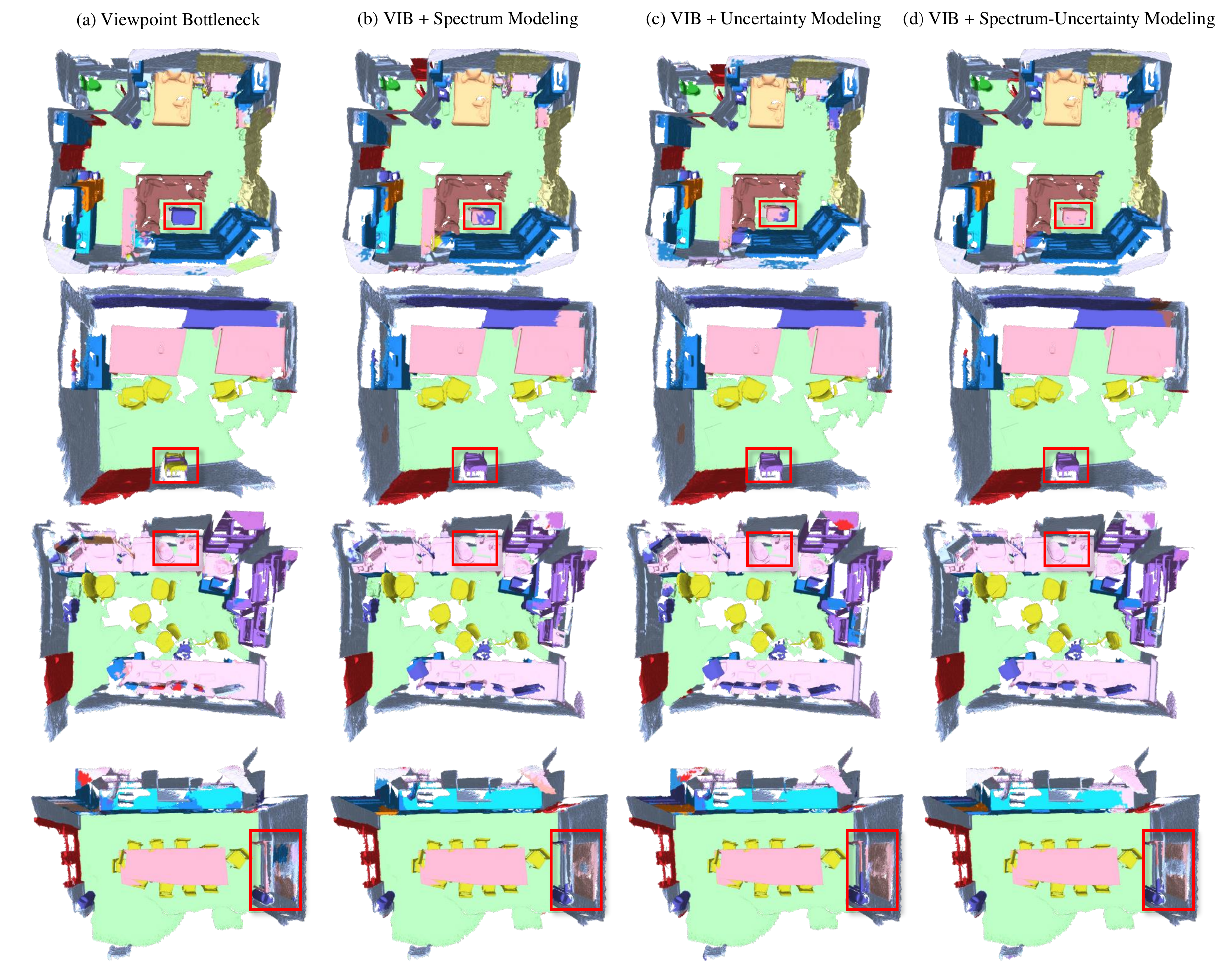}}
    \caption{The semantic segmentation qualitative results of Viewpoint Bottleneck, VIB with Spectrum Modeling, VIB with Uncertainty Modeling and VIBUS on the ScanNet validation set. The results are obtained with the setting of 200 inital annotations.}
    \label{fig:ablation_visualized}
\end{figure*}

We also provide visualization results of different modeling strategies in 200-point setting in  Fig. \ref{fig:ablation_visualized}.

\textbf{Downsample ratio in Uncertainty-Spectrum Modeling} Firstly, to evaluate how the downsampling process affects the performance of spectral clustering, we use different numbers of vertices as the target of $Quadric\ Edge$ $Collapse\ Decimation$ when before spectral clustering.
Table. \ref{tbl:ablation_dsrate} shows the clustering results are robust to the number of target vertices of spectral clustering.
We set this parameter to 8000 in other experiments.

\textbf{Embedding length in Uncertainty-Spectrum Modeling} Also, we test different embedding lengths when performing spectral clustering and report the mIoUs of clustering results in Table. \ref{tbl:ablation_featdim}.
With an embedding length of 50, the spectral clustering gives optimal results when provided with 20, 50, and 200 initial annotations.
In the setting of 100 initial annotations, the spectral clustering performs the best with an embedding length of 90 or 110.
In other experiments, we set the embedding length to 50.

\newpage
\textbf{Ratio of geodesic and angular distance} Table. \ref{tbl:ablation_gdad} reports the mIoUs of spectral clustering results with different percentage of geodesic distance ($\delta$) in the spectrum distance measure.
We observe that using angular distance only ($\delta=0$) and distance only ($\delta=1$) both give unsatisfying results.
With $\delta=0.4$ in the settings of 20, 50, 100 initial annotations and with $\delta=0.6$ in the settings of 100 and 200 initial annotations, spectral clustering has optimal performances.
In other experiments, we set $\delta$ to 0.6.

\subsection{Hardware Environment and Time Complexity}
VIBUS is an efficient large-scale 3D scene parsing method.
The experiments are conducted on a computing server which have 2 AMD EPYC 7742 CPUs (128 cores in total), 256GB memory and 8 Geforce RTX 3090 GPU Cards.
The server runs on Ubuntu 20.04.4 LTS operating system with linux kernel 5.13.0.

We use 4, 2 and 1 GPU cards in the pretraining, fine-tuning and inference stages respectively.
We list the computational overhead of pre-training, modeling, fine-tuning and inference on ScanNetV2 in Table. \ref{tbl:time}.

\section{Limitations and Future Works}

Despite the novelty of our proposed methods, there are still some limitations that can be improved in future works.

Firstly, our methods achieve unsatisfying results on small datasets like Semantic3D (which has only 9 scenes for training).
Without sufficient scenes, it is hard for the Viewpoint Bottleneck pretraining pipeline to learn how to maintain the information about the input point cloud against the random transformations.
In the future, we will seek for a solution that has smaller reliance on the size of the dataset.

Secondly, our proposed methods cannot handle the non-manifold meshes, in which case the calculation of the geodesic distances is inaccurate.
To solve this problem, we will try performing mesh completing algorithm, or find a more robust subsitute for calculating geodesic distances other than heat diffusion methods.

\begin{table}[htbp]
  \centering
  \caption{Data-efficient 3D Scene semantic segmentation Result of Different Modelling Approaches on ScanNet Validation Set with 100 Initial Labels (IoU: \%)}
  \resizebox{\linewidth}{!}{ 
    \begin{tabular}{cccccccccccccccccccccc}
    \toprule
    setting & category & Wall  & Floor & Cabinet & Bed   & Chair & Sofa  & Table & Door  & Window & B.S. & Picture & Counter & Desk  & Curtain & Fridge & S.C.  & Toilet & Sink  & Bathtub & O.F. \\
    \midrule
    \multirow{3}[2]{*}{200} & joint & 82.6  & 95.5  & 62.3  & 78.6  & 90.3  & 84.8  & 70.5  & 58.8  & 57.2  & 77.1  & 29.2  & 64.1  & 58.7  & 70.9  & 48.9  & 65.6  & 91.2  & 63.3  & 81.3  & 56.2 \\
          & uncertainty & 83.3  & 95.7  & 63.2  & 79.1  & 90.1  & 83.2  & 71.4  & 59.2  & 55.5  & 79.1  & 30.4  & 65.6  & 61.1  & 70.7  & 45.5  & 67.1  & 91.0    & 61.9  & 84.5  & 55.3 \\
          & spectrum & 83.5  & 95.7  & 62.8  & 80.3  & 89.4  & 82.0    & 70.5  & 58.6  & 55.7  & 78.0    & 31.2  & 62.5  & 61.8  & 67.8  & 46.5  & 63.7  & 90.6  & 62.1  & 85.6  & 52.8 \\
    \midrule
    \multirow{3}[2]{*}{100} & joint & 81.6  & 95.8  & 62.0    & 78.9  & 89.7  & 82.2  & 72.2  & 58.7  & 54.9  & 73.6  & 29.9  & 62.9  & 60.3  & 71.2  & 50.4  & 67.7  & 91.1  & 60.4  & 82.5  & 51.3 \\
          & uncertainty & 82.2  & 95.8  & 63.0    & 79.9  & 89.3  & 83.1  & 72.7  & 57.4  & 58.2  & 73.7  & 28.5  & 63.7  & 62.0    & 71.1  & 46.3  & 65.3  & 91.4  & 59.0    & 82.2  & 48.2 \\
          & spectrum & 82.7  & 95.8  & 62.4  & 81.2  & 89.0    & 84.0    & 71.8  & 59.5  & 55.0    & 75.7  & 27.4  & 61.7  & 60.4  & 72.7  & 47.4  & 66.7  & 89.4  & 56.5  & 84.0    & 47.7 \\
    \midrule
    \multirow{3}[2]{*}{50} & joint & 79.0    & 95.7  & 56.9  & 77.8  & 88.0    & 81.6  & 69.5  & 52.7  & 50.0    & 74.2  & 19.7  & 30.2  & 57.3  & 71.4  & 44.4  & 67.3  & 90.0    & 59.8  & 82.0    & 45.1 \\
          & uncertainty & 80.2  & 95.7  & 57.9  & 77.8  & 88.7  & 81.4  & 70.1  & 52.9  & 50.9  & 75.0    & 18.8  & 29.0    & 57.4  & 70.4  & 46.2  & 66.3  & 89.8  & 57.7  & 83.2  & 44.8 \\
          & spectrum & 80.6  & 95.6  & 59.2  & 78.9  & 87.8  & 82.5  & 68.2  & 52.6  & 48.7  & 75.5  & 18.0    & 61.0    & 56.2  & 69.8  & 42.2  & 63.7  & 87.2  & 57.4  & 83.9  & 43.5 \\
    \midrule
    \multirow{3}[2]{*}{20} & joint & 76.0    & 95.4  & 51.1  & 71.7  & 87.0    & 78.6  & 66.0    & 42.1  & 43.6  & 69.5  & 0.0     & 50.3  & 49.7  & 61.1  & 34.2  & 58.0    & 87.4  & 57.4  & 79.0    & 43.0 \\
          & uncertainty & 77.1  & 95.4  & 51.9  & 73.9  & 86.9  & 80.2  & 67.3  & 42.4  & 43.7  & 69.1  & 0.0     & 51.1  & 51.2  & 61.2  & 34.9  & 60.7  & 85.9  & 58.0    & 80.1  & 43.6 \\
          & spectrum & 77.9  & 95.4  & 52.3  & 74.9  & 85.8  & 78.7  & 66.0    & 41.7  & 44.8  & 70.8  & 13.8  & 51.5  & 50.5  & 61.6  & 33.6  & 56.9  & 84.4  & 57.0    & 79.8  & 43.4 \\
    \midrule
    \multicolumn{22}{l}{* The B.S. is the Bookshelf, the S.C. is the shower curtain and the O.F. is the other furniture.} \\
    \end{tabular}%
    }
  \label{tab:class_iou}%
\end{table}%

\section{Conclusion}
\label{sec:conclusion}

In this work, we address the problem of data-efficient 3D scene parsing and propose a principled method named VIBUS with two technical modules: 1) Viewpoint Bottleneck, a self-representation learning pipeline that has simple network structures and avoids selecting sample pairs, and 2) Uncertainty-Spectrum modeling, an effective approach to harvest reliable pseudo labels by filtering a set of generated label. With simple implementation, our method achieves large margins over existing methods on the ScanNetV2 dataset and wins the leading place on the online server.

\bibliography{elsarticle-template}

\end{document}